\documentclass[12pt, a4paper]{article}
\usepackage{xr}
\usepackage{array,bm,amsmath,amssymb,mathrsfs,epsfig,epstopdf}
\usepackage{amsthm}
\usepackage{setspace}
\usepackage{CJK}
\usepackage{color}
\usepackage{lscape}
\usepackage{verbatim}
\usepackage{enumerate}
\usepackage{booktabs}
\usepackage{threeparttable}
\usepackage{multicol}
\usepackage{longtable}
\usepackage{multirow}
\usepackage{bm}
\usepackage{bbm}
\usepackage{cases}
\usepackage{caption}
\usepackage{natbib}
\usepackage{algorithm}
\usepackage{algorithmic}
\usepackage{geometry}
\usepackage{framed}
\usepackage[pdftex,bookmarksnumbered,bookmarksopen,colorlinks, citecolor=blue,linkcolor=blue,urlcolor=blue]{hyperref}
\usepackage{graphicx}
\usepackage{amsmath}
\usepackage{amsfonts}
\usepackage{float}
\usepackage{ragged2e}
\usepackage[figuresright]{rotating}
\usepackage{mathtools}
\usepackage{pifont}
\usepackage{indentfirst}
\usepackage{listings}
\definecolor{codegreen}{rgb}{0,0.6,0}
\definecolor{codegray}{rgb}{0.5,0.5,0.5}
\definecolor{codepurple}{rgb}{0.58,0,0.82}
\definecolor{backcolour}{rgb}{0.95,0.95,0.92}
\lstdefinestyle{mystyle}{
	backgroundcolor=\color{backcolour},   
	commentstyle=\color{codegreen},
	keywordstyle=\color{magenta},
	numberstyle=\tiny\color{codegray},
	stringstyle=\color{codepurple},
	basicstyle=\ttfamily\footnotesize,
	breakatwhitespace=false,         
	breaklines=true,                 
	captionpos=b,                    
	keepspaces=true,                 
	numbers=left,                    
	numbersep=5pt,                  
	showspaces=false,                
	showstringspaces=false,
	showtabs=false,                  
	tabsize=2
}

\newcommand\gA{{\mathcal{A}}}
\newcommand\gB{{\mathcal{B}}}
\newcommand\gC{{\mathcal{C}}}

\newcommand\gF{{\mathcal{F}}}
\newcommand\gG{{\mathcal{G}}}

\newcommand\gN{{\mathcal{N}}}
\newcommand\gO{{\mathcal{O}}}

\newcommand\gR{{\mathcal{R}}}
\newcommand\gS{{\mathcal{S}}}

\newcommand\gY{{\mathcal{Y}}}
\newcommand\gZ{{\mathcal{Z}}}

\DeclareMathOperator*{\argmax}{arg\,max}
\DeclareMathOperator*{\argmin}{arg\,min}

\newtheorem{theorem}{Theorem}[section]
\newtheorem{lemma}{Lemma}[section]
\newtheorem{Definition}{Definition}[section]
\newcommand{\tabincell}[2]{\begin{tabular}{@{}#1@{}}#2\end{tabular}}

\title{\bf Interval-Valued Time Series Classification Using \(D_K\)-Distance}
\author{Wan Tian$^1$, Zhongfeng Qin$^{1,2}$\thanks{Corresponding author}}
\date{ wantian61@foxmail.com, qin@buaa.edu.cn\\
	$^1$ School of Economics and Management, Beihang University, Beijing 100191, China \\%
	$^2$Key Laboratory of Complex System Analysis, Management and Decision (Beihang University), Ministry of Education, Beijing 100191, China \\[2ex]%
}

\begin{document}
\maketitle

\begin{abstract}
In recent years, modeling and analysis of interval-valued time series have garnered increasing attention in econometrics, finance, and statistics. However, these studies have predominantly focused on statistical inference in the forecasting of univariate and multivariate interval-valued time series, overlooking another important aspect: classification. In this paper, we introduce a classification approach that treats intervals as unified entities, applicable to both univariate and multivariate interval-valued time series. Specifically, we first extend the point-valued time series imaging methods to interval-valued scenarios using the \(D_K\)-distance, enabling the imaging of interval-valued time series. Then, we employ suitable deep learning model for classification on the obtained imaging dataset, aiming to achieve classification for interval-valued time series. 	
In theory, we derived a sharper excess risk bound for deep multiclassifiers based on offset Rademacher complexity. Finally, we validate the superiority of the proposed method through comparisons with various existing point-valued time series classification methods in both simulation studies and real data applications.

\end{abstract}

\maketitle

\section{Introduction} \label{sec1}
Interval-valued time series, capable of simultaneously capturing the level and variation information of the underlying data generating process, have gained increasing attention in the fields of finance, econometrics, and statistics in recent years \citep{gonzalez2013constrained, han2016vector, sun2018threshold, sun2022model}. Both its theoretical foundations and practical applications have experienced significant development \citep{gonzalez2013constrained, yang2016analysis, sun2019asymmetric, he2021forecasting,  lu2022forecasting}. For instance, in the medical field, observations of diastolic and systolic blood pressure within a monitoring period; in meteorology, the daily extremes of air indicators such as temperature, humidity, and pollutant concentration; and in finance, the bid and ask prices within a trading period, inflation rates, short-term, and long-term interest rates all constitute interval-valued time series. Compared to conventional point-valued time series, modeling interval-valued time series has two distinct advantages. First, within the same observation period, interval-valued observations contain more variation and level information compared to point-valued observations \citep{han2012autoregressive, han2016vector}. This implies that modeling based on interval-valued data leads to more efficient estimations and powerful inferences. Second, while specific disturbances might disastrously affect inferences based on point-valued data, this issue can be addressed through modeling interval-valued data \citep{han2016vector}, indicating that modeling interval-valued data is more robust.

Over the past three decades, numerous methods for analyzing interval-valued time series have been continually proposed from both theoretical and applied perspectives \citep{arroyo2007exponential, maia2008forecasting, arroyo2011different, gonzalez2013constrained, han2012autoregressive, han2016vector, wanginterval}, yet almost all have been primarily focused on forecasting. \citet{arroyo2007exponential} utilizes bivariate point-valued time series to represent interval-valued time series and extends classical exponential smoothing methods to interval scenarios with the aid of interval arithmetic, comparing its performance with interval multilayer perceptrons. \citet{maia2008forecasting} represent intervals using center and range measures. Then, they employ autoregressive, autoregressive integrated moving average, and artificial neural network separately for forecasting, followed by a combination of these models to enhance forecasting efficiency. With the help of interval arithmetic, \citet{arroyo2011different} represents interval-valued time series using various methods and compares the performance of multiple classical and statistical learning methods on financial interval-valued time series.
The methods mentioned earlier essentially predict point-valued time series to achieve forecasting for interval-valued time series. \citet{gonzalez2013constrained} proposes an effective modeling approach for interval-valued time series considering constraints related to the size of upper and lower bounds. It presents a two-stage efficient parameter estimation method based on conditional log-likelihood functions and the corresponding statistical theory. \citet{han2012autoregressive} introduces the concept of extended random intervals, disregarding the relationship between upper and lower bounds, and establishes the autoregressive conditional interval model. Leveraging this concept,  \citet{han2016vector} develops a vector autoregressive moving average model for interval vector-valued time series, constructing corresponding estimation methods and statistical theory.
A comprehensive review of interval-valued data research can be found in \citet{wanginterval}, which systematically compares and analyzes various definitions of interval-valued time series, categorizes research within established frameworks, and identifies potential future research directions.

Within the framework of supervised learning, classification and forecasting are two equally important subtasks. However, to our knowledge, research on classification methods for interval-valued data, whether approached from point-wise representation or an overall interval perspective, is extremely limited. Moreover, addressing interval-valued time series that require consideration of dependencies is even scarcer. \citet{palumbo1999non} extends factorial classification analysis to interval-valued data and introduces a three-stage classification procedure. \citet{rasson2000symbolic} generalizes the Bayesian classification analysis method to interval-valued data and provides interval prior and posterior estimation methods based on kernel density estimation. \citet{qi2020interval} introduces a unified representation framework for interval-valued data and applies common ensemble learning methods for classification purposes. However, these classification methods are not specifically designed for interval-valued time series. Recently, \citet{wan2023discriminant} proposed a  adaptive classification method applicable to both univariate and multivariate interval-valued time series, providing corresponding bounds for generalization errors. This method employs convex combinations of upper and lower bounds for classification, and treating the convex combination coefficients as learnable parameters.

Classification methods for both univariate and multivariate point-valued time series have witnessed significant advancements in recent years due to the continuous evolution of deep learning \citep{bagnall2017great, ismail2019deep, ruiz2021great}. These methods have demonstrated immense potential in various benchmark datasets and real-world applications \citet{bagnall2017great}, broadly categorized into five classes: distance-based methods \citep{lines2015time, cuturi2017soft, lucas2019proximity}, dictionary-based methods \citep{schafer2015boss, schafer2016scalable, schafer2017fast}, Shapelets-based methods \citep{schafer2017fast}, ensemble-based methods \citep{bagnall2015time, lucas2019proximity}, and deep learning-based methods \citep{Wang2016TimeSC, Cui2016MultiScaleCN, ismail2019deep}. However, these methods are neither directly applicable to interval-valued time series nor have they been effectively extended to cater to interval-valued time series data.

In this paper, we propose an effective classification method applicable to both univariate and multivariate interval-valued time series, treating interval as a unified entity. Specifically, we extend the imaging methods of point-valued time series to interval-valued scenarios using the \(D_K\)-distance. The imaging methods corresponding to univariate and multivariate point-valued time series are the Recurrence Plot (RP) \citep{eckmann1995recurrence} and Joint Recurrence Plot (JRP) \citep{ROMANO2004214}, respectively. Then, leveraging these extended imaging methods, we convert interval-valued time series into images. Then, we select suitable deep learning model to classify the obtained image dataset, aiming to achieve classification of interval-valued time series. Theoretically, we obtained the optimal convergence rate of excess risk for deep multi-classification based on softmax using the offset Rademacher complexity proposed by \citet{liang2015learning}. We validated the effectiveness of our proposed method through simulation studies and weather data from China.
In addition, we use various types of point-valued time series methods to classify several representative points within intervals and comparing.	

The structure of the paper is as follows: Section \ref{sec2} introduces the \(D_K\)-distance and extends point-valued time series imaging methods to interval-valued scenarios accordingly. Section \ref{sec3} presents the convergence rate of excess risk for multiclass classification based on offset Rademacher complexity. Sections \ref{sec4} and \ref{sec5} provide simulation studies and real data applications, respectively. Section \ref{sec6} concludes the paper. All technical proofs are included in the Appendix \ref{appendix}.

\section{Methodology} \label{sec2}
In this section, we introduce the \(D_K\)-distance and discuss the impact of kernel function selection on its computational aspects. Then, based on the \(D_K\)-distance, we extend the univariate and multivariate time series imaging methods RP and JRP to the interval-valued scenario. Finally, we formalize the classification of interval-valued time series as an image classification problem.

\subsection{The \(D_K\)-distance} \label{sec21}
Let \( A = [A^l, A^u] \) and \( B = [B^l, B^u] \) be two intervals, where \( A^u \) and \( A^l \) represent the upper and lower bounds, respectively. Alternatively, we can express the intervals as \( A = (A^c, A^r) \) and \( B = (B^c, B^r) \), where \( A^c = (A^l + A^u)/2 \) and \( A^r = (A^u - A^l)/2 \) represent the center and range of the interval, respectively. The same applies to interval \( B \). The \(D_K\)-distance serves as a metric for measuring the distance between pairs of intervals by considering the absolute differences between all possible pairs of points from intervals, which is defined as
\begin{equation} \label{dkdis}
	D_K(A, B) = \sqrt{\int_{u, v\in \mathbb{S}^0}[s_A(u) - s_B(u)][s_A(v) - s_B(v)]d K(u, v)},
\end{equation}
where \(\mathbb{S}^0 = \{u \in \mathbb{R}, |u| = 1\}\) is the unit sphere in \(\mathbb{R}\), \(s_A(u)\) and \(s_B(u)\) are respectively the support function of interval \(A\) and \(B\), defined as
\[
s_A(u) = \sup_{a\in A} \langle u,a \rangle, \ s_B(u) = \sup_{b\in B} \langle u,b \rangle, \  u \in \mathbb{S}^0,
\]
where \(\langle \cdot,\cdot \rangle\) represents the inner product. In the univariate interval-valued scenario, the unit sphere \(\mathbb{S}^0 = \{-1, 1\}\), and \(s_A(u) = A^u\) if \(u = 1\), \(s_A(u) = -A^l\) if \(u = -1\). The kernel \(K(u, v)\) is a symmetric positive definite function for \(u,v \in \mathbb{S}^0\), satisfying
\begin{equation} \label{kernelcondition}
\begin{cases}
	&K(1, 1) >0, \\
	&K(1, 1) K(-1, -1) > K^2(1, -1), \\
	&	K(1, -1) =K(-1, 1).
\end{cases}
\end{equation}

Based on the definitions of the support functions \(s_A(u)\) and \(s_B(u)\), the \(D_K\)-distance can be equivalently represented in the following quadratic form,
\begin{equation} \label{dkqua}
\begin{aligned}
D^2_K(A, B) &= K(1,1) (A^u - B^u)^2 + K(-1, -1) (A^l - B^l)^2 - 2K(1, -1)(A^l - B^l)(A^u - B^u)\\
&= \left[
\begin{array}{c}
A^u - B^u\\
-(A^l - B^l)
\end{array}
\right]^\top \left[
\begin{array}{cc}
K(1, 1) & K(1, -1)\\
K(-1, 1) & K(-1, -1)\\
\end{array}
\right]  \left[
\begin{array}{c}
A^u - B^u\\
-(A^l - B^l)
\end{array}
\right].
\end{aligned}
\end{equation}

Below, utilizing the quadratic form representation, we discuss the impact of kernel function selection on the computational aspect of the \(D_K\)-distance. Note that the kernel we select does not necessarily have to be positive definite. As a first example, the form of the kernel function we select is
\[
K = \left[
\begin{array}{cc}
	K(1, 1) & K(1, -1)\\
	K(-1, 1) & K(-1, -1)\\
\end{array}
\right] = 
\left[
\begin{array}{cc}
	1/4 & - 1/4\\
	- 1/4 & 1/4 \\
\end{array}
\right], 
\]
the resulting expression for the \(D_K\)-distance as \((A^c - B^c)^2\), which implies that the center serves as a representative point for measuring the distance between pairwise intervals.

As a second example, with the selected kernel as 
\[
K = \left[
\begin{array}{cc}
	K(1, 1) & K(1, -1)\\
	K(-1, 1) & K(-1, -1)\\
\end{array}
\right] = 
\left[
\begin{array}{cc}
	1 & 1\\
	1 & 1\\
\end{array}
\right], 
\]
this results in \(D^2_K(A,B) = 4(A^r -B^r)^2\), implying that the range serves as the representative point for the intervals. As a third example, with the selected kernel satisfying \(K(1, 1) = K(-1, -1)\) and \(|K(-1, 1)| < K(-1, -1)\), we obtain 
\[
D^2_K(A, B) = 2(K(1, 1)+ K(-1, 1))(A^r-B^r)^2 + 2(K(1, 1) - K(-1, 1))(A^c - B^c)^2, 
\] 
which implies that both the center and the range pairs serve as representative points for the intervals. As a fourth example, setting \(K(-1, 1) = 0\) immediately gives us \(D^2_K(A, B)= K(1,1) (A^u - B^u)^2 + K(-1, -1) (A^l - B^l)^2\). This signifies that the upper and lower bounds of the intervals act as representative points to measure the distance between pairwise intervals.

The four cases above illustrate that different kernels utilize different representative points when calculating the \(D_K\)-distance, resulting in varying interval information used for subsequent classification. As our research subject, interval-valued time series consist of interval-valued random variables indexed by time. For two interval-valued time series \(X_1 = \{X_{1,t}\}^T_{t=1}\) and \(X_2 = \{X_{2, t}\}^T_{t=1}\) of length \(T\), their \(D_K\)-distance is defined as
\[
D^2_K \left(X_1, X_2\right)= \sum_{t=1}^{T} D^2_K\left(X_{1,t}, X_{2,t}\right). 
\]

\subsection{Interval recurrence plot} \label{sec22}
Given a dataset \(\gS_{1,n} = \{(X_i,Y_i)\}^n_{i=1}\) of univariate interval-valued time series used for classification, where \(X_i = \{X_{i, t}\}^T_{t=1}\) represents univariate interval-valued time series with length \(T\), and \(Y_i \in \gY =\{1, 2,\cdots, C\}\) is its corresponding label, where \(C\) is the maximum number of categories. Below, we extend the univariate point-valued time series imaging method RP to interval-valued scenarios using the \(D_K\)-distance discussed in Section \ref{sec21}. Similar to RP, for any interval-valued time series observation \(X_i, i = 1,2,\cdots, n\), we first extract trajectories of length \(m\) with time delay \(\kappa\),
\begin{equation}\label{trajectories}
\vec{X}_{i, j} = (X_{i, j}, X_{i, j+\kappa}, \cdots, X_{i, j+(m-1)\kappa}) , j = 1,2,\cdots, T-(m-1)\kappa,  
\end{equation}
where each element in the trajectory is an interval instead of a point. Replacing the Euclidean norm in RP with the \(D_K\)-distance, we have the distance between pairwise trajectories as  
\begin{equation}\label{trajectoriesdis}
I_i(j, k) = H(\epsilon_i -  D_K(\vec{X}_{i,j}, \vec{X}_{i, k})),\ j,k = 1,2,\cdots, T - (m-1)\kappa,
\end{equation}
where \(H(\cdot)\) is a Heaviside step function, and \(\epsilon_i\) is a threshold that can vary for different observations. The Heaviside step function is predefined in the Recurrence Plot, and we adopt this function. It is worth noting that \(m\), \(\kappa\), and \(\epsilon_i, i = 1,2,\cdots, n\) are tuning parameters, which are determined in subsequent experiments. The two steps consisting of (\ref{trajectories}) and (\ref{trajectoriesdis}), which transform interval-valued time series into images, are referred to as the Interval Recurrence Plot (IRP). By utilizing IRP, we can transform each observation \(X_i\) into an image \(I_i = (I_i(j, k))_{1\leq j,k \leq T-(m-1)\kappa}\). Therefore, we can transform the interval-valued time series dataset \(\mathcal{S}_{1,n}\) into an image dataset \(\mathcal{D}_{1,n} = \{(I_i,Y_i)\}^n_{i=1}\). We only need to classify \(\mathcal{D}_{1,n}\) to achieve the purpose of classifying \(\mathcal{S}_{1,n}\). At this point, we can fully utilize some advanced deep learning models to enhance the classification performance of the image dataset \(\mathcal{D}_{1,n}\). Algorithm \ref{IRP} outlines the process of transforming \(\gS_{1,n}\) into the image dataset \(\mathcal{D}_{1,n}\).
\begin{algorithm}[H]
\caption{Using IRP for imaging \(\gS_{1,n}\).} \label{IRP}
\begin{algorithmic}
\REQUIRE ~~\\
Dataset \(\gS_{1,n}\), length \(m\), time delay \(\kappa\), threshold \(\epsilon_1, \epsilon_2, \cdots, \epsilon_n\), kernel \(K\)
\ENSURE ~~\\
for \(i = 1,2,\cdots, n\):\\ 
\quad using (\ref{trajectories}) to extract \(T - (m-1)\kappa\) trajectories;\\
\quad for \(j = 1,2,\cdots, T - (m-1)\kappa\):\\
\quad \quad for \(k = 1,2,\cdots, T - (m-1)\kappa\):\\
\quad \quad\quad using (\ref{trajectoriesdis}) to obtain distances \(I_i(j, k) \) between trajectories \(j\) and \(k\);\\
\textbf{Output:} \(\mathcal{D}_{1,n} = \{(I_i, Y_i)\}^n_{i=1}\).\\
\end{algorithmic}
\end{algorithm}

\subsection{Interval joint recurrence plot} \label{sec23}
Given a multivariate interval-valued time series dataset \(\gS_{2,n} = \{(W_i, Y_i)\}^n_{i=1}\) for classification, where \(W_i= (W_{i, j, t})_{1\leq j \leq d, 1\leq t\leq T}\) represents a \(d\)-dimensional interval-valued time series of length \(T\), and \(Y_i \in \gY = \{1,2,\cdots, C\}\) is its corresponding label. Below, we extend the multivariate point-valued time series imaging method JRP \citep{ROMANO2004214} to interval-valued scenarios using the \(D_K\)-distance discussed in Section \ref{sec21}.

The multivariate point-valued time series imaging method JRP initially employs RP to convert each dimension of the multivariate point-valued time series into images. Similarly, for multivariate interval-valued time series, we first utilize the IRP proposed in Section \ref{sec22} to transform each dimension of the multivariate interval-valued time series into images. Specifically, we first extract the trajectories of each dimension of the multivariate interval-valued time series, and then compute the distance between trajectories, defined as
\begin{equation}\label{dimendis}
I_{i,j}(k, w) = H(\epsilon_{i,j} - D_K(\vec{W}_{i,j, k}, \vec{W}_{i,j, w})), \ i = 1,2,\cdots,n, j = 1,2,\cdots, d,
\end{equation}
where \(\vec{W}_{i,j, k} = (W_{i,j,k}, W_{i,j,k+\kappa}, \cdots, W_{i,j,k + (m-1)\kappa}), \vec{W}_{i,j, w} = (W_{i,j,w}, W_{i,j,w+\kappa}, \cdots, W_{i,j,w + (m-1)\kappa})\), \(k, w = 1,2,\cdots, T-(m-1)\kappa\). For any multivariate interval-valued time series observation \(W_i\), we can obtain \(d\) recurrence plots \(I_{i,1}, I_{i,2}, \cdots, I_{i,d}\). The second step of JRP is to aggregate the information of each dimension using the Hadamard product (element-wise). Similarly, we also utilize the Hadamard product to aggregate the information of each dimension of the multivariate interval-valued time series, and thereby obtain the final image, defined as
\begin{equation} \label{aggregatinginformation}
I_i = I_{i,1} \circ I_{i,2} \circ \cdots \circ I_{i,d},
\end{equation}
where \(I_{i,j} = (I_{i,j}(k, w))_{1\leq k,w \leq T-(m-1)\kappa}, j = 1,2,\cdots, d\), \(\circ\) denotes the Hadamard product. The two preceding steps are referred to as Interval Joint Recurrence Plots (IJRP). With IJRP, we can transform a multivariate interval-valued time series dataset \(\gS_{2,n}\) into an image dataset \(\mathcal{D}_{2,n}= \{(I_i, Y_i)\}^n_{i=1}\). Then, we only need to classify \(\mathcal{D}_{2,n}\) to achieve the purpose of classifying \(\gS_{2,n}\). 

\begin{algorithm}[H]
\caption{Using IJRP for imaging \(\gS_{2,n}\).} \label{IJRP}
\begin{algorithmic}
\REQUIRE ~~\\
Dataset \(\gS_{2,n}\), length \(m\), time delay \(\kappa\), threshold \(\epsilon_{1, 1}, \epsilon_{1, 2}, \cdots, \epsilon_{n, d}\), kernel \(K\).
\ENSURE ~~\\
for \(i = 1,2,\cdots, n\):\\ 
\quad for \(j = 1,2,\cdots, d\):\\ 
\quad\quad using (\ref{trajectories}) to extract \(T - (m-1)\kappa\) trajectories;\\
\quad\quad for \(k = 1,2,\cdots, T - (m-1)\kappa\):\\
\quad\quad \quad for \(w = 1,2,\cdots, T - (m-1)\kappa\):\\
\quad\quad \quad\quad using (\ref{dimendis}) to obtain distances \(I_{i,j}(k, w) \) between trajectories \(k\) and \(w\);\\
\quad using (\ref{aggregatinginformation}) to aggregating information.\\
\textbf{Output:} \(\mathcal{D}_{2,n}= \{(I_i, Y_i)\}^n_{i=1}\).\\
\end{algorithmic}
\end{algorithm}

\subsection{Classification based on imaging datasets}\label{sec24}
In this section, we formalize the classification problem of interval-valued time series as an image classification problem. Utilizing the imaging methods IRP and IJRP proposed in Sections \ref{sec22} and \ref{sec23}, respectively, we can transform the interval-valued time series data into an image dataset, thereby converting the classification problem of interval-valued time series into an image classification problem. 

Without loss of generality, we assume that the image dataset we need to classify is denoted by \(\mathcal{D}_n = \{(I_i, Y_i)\}^n_{i=1}\), where it can be either \(\mathcal{D}_{1,n}\) or \(\mathcal{D}_{2,n}\). The deep learning model used for classification is denoted by \(F(\cdot, \Theta)\), with \(\Theta\) represents the parameters of the model. The performance of classification is measured using the loss \(L_F(\cdot, \cdot)\), where a smaller value indicates better classification performance. Therefore, the optimization objective is
\begin{equation} \label{allopt}
\min_\Theta \frac{1}{n}\sum_{i=1}^{n}L_F(F(I_i, \Theta), Y_i).
\end{equation}

\begin{figure}[H]
\centering
\includegraphics[scale=0.66]{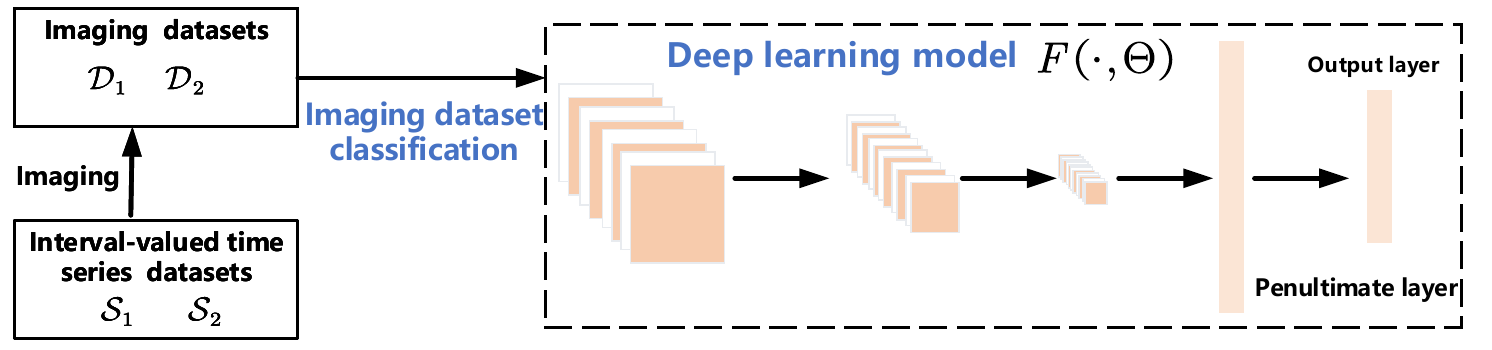}
\caption{The technical roadmap of classifying interval-valued time series by transforming them into images.}\label{technicalroadmap}
\end{figure}

In general, we can utilize method like stochastic gradient descent (SGD) to optimize objective (\ref{allopt}), obtaining estimates for the model parameters \(\Theta\). Figure \ref{technicalroadmap} illustrates the technical roadmap of the paper.

\section{Excess risk bound}\label{sec3}
Excess risk is an important concept in statistics and machine learning, as it reflects the generalization ability of model on unseen data. With the increasing popularity of deep learning, there has been a growing body of research on the generalization theory of neural networks \citep{zhang2023mathematical, jentzen2023mathematical}. These studies can be roughly divided into two categories based on whether they consider the network structure when constructing excess risk bounds \citep{he2020recent, bartlett2021deep, duan2023fast}. In this section, we establish the optimal convergence rate of excess risk for deep multi-classifiers based on offset Rademacher complexity \citep{liang2015learning} , ignoring the structure of neural networks.

Assuming that by minimizing the optimization objective (\ref{allopt}) using certain optimization methods, we can obtain an estimation \(\widehat{\Theta}\) for the model parameters \(\Theta\), i.e.,
\[
\widehat{\Theta} \in \argmin_\Theta \frac{1}{n}\sum_{i=1}^{n}L_F(F(I_i, \Theta), Y_i).
\]

Then, the input layer to the penultimate layer of the deep learning model \(F(\cdot, \widehat{\Theta})\) acts as a deep feature extractor for the input data, denoted as \(\psi(\cdot)\), which transforms images into vectors in Euclidean space, namely
\[
Z_i = \psi(I_i) \in \mathbb{R}^p, \ i = 1,2,\cdots, n,
\]
where \(I_i\) originate from \(\mathcal{D}_n\), \(Z_i\) is the penultimate layer of the deep learning model, commonly referred to as a feature map. Then, we only need to classify the dataset \(\gS_n = \{(Z_i, Y_i)\}^n_{i=1}\) to achieve the classification of the original interval-valued time series. It is important to note that during the training process of deep learning model, the mapping \(\psi(\cdot)\) and the final layer of the linear classifier is determined concurrently. 

We construct excess risk bounds for the multi-classification task based on dataset \(\gS_n\). We first introduce some necessary notations. Let \((Z, Y)\) conform to an unknown joint measure \(P\) on \(\gZ \times \gY\), where \(\gZ \subseteq \mathbb{R}^p\), and \(\gY = \{1,2, \cdots, C\}\) with \(C \geq 2\). This is a multi-classification problem, and naturally, subsequent empirical analyses also involve multi-classification. Let \(\gF = \{f: \gZ \times \gY \to \mathbb{R}\}\) denote the hypothesis space for the classification function \(f\), and we will provide the specific form of \(\gF\) later. In multi-classification, for a given \(f\), the predicted label corresponding to \(Z \in \gZ \) is \(\widehat{Y} = \argmax_{Y \in \gY} f(Z, Y)\), i.e., the one with the highest score. 

Our aim is to select a function from the hypothesis space \(\gF\) that accurately predicts the labels of \(Z \in \gZ\). Due to the non-convexity of the 0-1 loss and the fact that its corresponding empirical risk is an NP-hard problem, which is computationally infeasible, we consider a surrogate loss. Note that there is a substantial body of research establishing the relationship between the 0-1 loss and surrogate losses corresponding to excess risk, as discussed in \citet{awasthi2022multi}, particularly addressing multi-classification scenarios. This paper exclusively establishes the convergence rate of excess risk corresponding to surrogate losses.

The current three commonly used multi-classification surrogate losses are \emph{max losses}, \emph{sum losses} and \emph{constrained losses}, each parameterized by an auxiliary function \(L(\cdot)\), which is non-negative and non-increasing on \(\mathbb{R}\). The common auxiliary functions include hinge loss \(\max(0, 1-a)\), squared hinge loss \(\max(0, 1-a)^2\), and exponential loss \(\exp(-a)\). In this paper, we focus on max losses \(\phi: \gF \times \gZ \times \gY \to [0, \infty )\). For \(f\in \gF\), \((Z, Y) \in \gZ \times \gY\), and auxiliary function \(L(\cdot)\), the max losses \(\phi(\cdot)\) is defined as: 
\[
\phi(f, Z, Y) = \max_{Y^\prime \neq Y} L(f(Z, Y) - f(Z, Y^\prime)) = L(f(Z, Y) - \max_{Y^\prime \neq Y} f(Z, Y^\prime)), 
\]
where the second equation holds because the auxiliary function is non-increasing, and \(f(Z, Y) - \max_{Y^\prime \neq Y} f(Z, Y^\prime)\) represents the so-called margin of a hypothesis \(f \in \gF\) for \((Z, Y)\).

Under the aforementioned notations, the \(\phi\)-risk is defined as \(R_\phi(f) = \mathbb{E}(\phi(f, Z, Y) )\), with its empirical counterpart being \(\widehat{R}_\phi(f) = \frac{1}{n} \sum_{i=1}^{n}\phi(f, Z_i, Y_i)\). Let
\[
\widehat{f}_\phi \in \argmin_{f \in \gF} \widehat{R}_\phi(f)
\]
be the empirical \(\phi\)-risk minimizer. Let \(R^*_\phi = \inf_f R_\phi(f)\) represent the optimal \(\phi\)-risk, and \(f^*_\phi\) be the corresponding minimizer. Deriving the convergence rate of the excess \(\phi\)-risk \(R_\phi(\widehat{f}_\phi) - R^*_\phi\), namely providing a generalization analysis of empirical \(\phi\)-risk minimizer \(\widehat{f}_\phi\), holds significant importance in statistics and machine learning.

The classical approach transforms the bounding of excess \(\phi\)-risk into bounding the global Rademacher complexity. Then, leveraging the complexity \(\gC(\gF)\) of the hypothesis space \(\gF\), such as covering numbers and VC dimension, bounds the global Rademacher complexity \citep{geer2000empirical, van2000asymptotic, gine2021mathematical, vaart2023empirical}. However, this classical method only yields suboptimal convergence rates \(\gO(\sqrt{\gC(\gF) /n})\). To address this issue, \citet{bartlett2005local} and \citet{koltchinskii2006local} proposed the so-called local Rademacher complexity, which primarily leverages the local structure of the hypothesis space \(\gF\) and can achieve optimal convergence rates \(\gO(\gC(\gF) /n)\). However, this improvement requires what is known as the Bernstein condition, which limits its applicability.

Recently, \citet{duan2023fast} has developed, under very mild conditions, the convergence rates of excess \(\phi\)-risk based on the offset Rademacher complexity proposed by \citet{liang2015learning} and applied it to several classes of parametric and non-parametric models. However, they only considered binary classification.  In this paper, we utilize offset Rademacher complexity to construct convergence rates of excess \(\phi\)-risk in multi-classification scenarios. To our knowledge, there is currently no related research available on this topic.

The definition of the offset Rademacher complexity proposed by \citet{liang2015learning}, denoted as 
\[
\gR^{\text{off}}_n(\gF, \varrho) = \mathbb{E}\left(\sup_{f\in \gF} \frac{1}{n} \sum_{i=1}^{n} \tau_i f(X_i) - \varrho f^2(X_i)
\right), \ \varrho >0,
\]
where \(\tau_1, \tau_2,\cdots, \tau_n\) represent independently and identically distributed Rademacher random variables. It is evident that \(\gR^{\text{off}}_n(\gF, \varrho)\) is a penalized version of the global Rademacher complexity. Let \(g(Z, f) = \mathbb{E}(\phi(f, Z, Y) - \phi(f^*_\phi, Z, Y)|Z)\) be the conditional excess \(\phi\)-risk, where it is evident that \(R_\phi(f) - R^*_\phi = \mathbb{E}(g(Z, f))\). In this paper, we focus on linear hypotheses, i.e.,
\[
\gF = \{f: (Z, Y) \to A_Y^\top Z + B_Y, A_Y \in \gA, B_Y \in \gB\}, 
\]
where \(\gA\) and \(\gB\) represent the parameter spaces for coefficients and biases, respectively. While there has been research considering nonlinear hypothesis spaces, such as single-layer ReLU networks \citep{awasthi2022multi}, our consideration of linear hypotheses aligns with the softmax utilized by CNNs for multi-classification. We assume the following standard conditions hold. The notation \( \lVert \cdot \rVert \) represents the spectral norm for matrices and the magnitude for vectors.

\emph{Condition } 1. There exist uniform constants \(c_A\) and \(c_B\) such that the coefficient space and bias space are upper bounded, i.e., \(\sup_{A\in \gA} \lVert A\rVert \leq c_A, \sup_{B\in \gB} \lVert B\rVert \leq c_B\).

\emph{Condition } 2. The feature mapping \(Z = \psi(I)\) obtained from any image is upper bounded by a constant \(c_Z\), i.e., \(\sup_{Z\in \gZ} \lVert Z\rVert \leq c_Z\).
 
Conditions 1 and 2 are standard. Building upon Condition 1, \citet{duan2023fast} and \citet{awasthi2022multi} separately investigated the excess risk bounds and the consistency of the hypothesis space. Condition 2 is intuitive; if the representation of an image is unbounded, it becomes impractical for downstream tasks in image processing. Given the above conditions, we can conclude the following regarding the function \(g(Z, f)\).

\begin{lemma}
Suppose Conditions 1 and 2 hold, then for any \(f_1, f_2 \in \gF\), we have
\[
|g(Z, f_1)| \leq 2\ell (c_Ac_Z + c_B), \ |g(Z, f_1) - g(Z, f_2)| \leq 4\ell (c_Ac_Z + c_B),
\]
where \(\ell\) is the Lipschitz constant of the auxiliary function \(L(\cdot)\). 
\end{lemma}

The auxiliary functions, namely hinge loss, squared hinge loss, and exponential loss, are all Lipschitz continuous, with Lipschitz constants of 1, 2, and 1, respectively. \citet{duan2023fast} assumes \(g(Z, f)\) to be not only Lipschitz continuous but also upper bounded when deriving the convergence rate of the excess \(\phi\)-risk for deep binary classification. However, in this paper, we only require the latter bounded condition.

\begin{theorem}\label{theorem31}
Suppose conditions 1 and 2 hold, the excess \(\phi\)-risk for multi-classification satisfies
\[
\mathbb{E}_{\gS_n}(R_\phi(\widehat{f}_\phi) - R^*_\phi) \leq  4 \mathcal{R}^{\text{off}}_n\left(\gG, \frac{1}{4\ell (c_Ac_Z + c_B)}\right).
\]
\end{theorem}

Theorem \ref{theorem31} tells us that the excess \(\phi\)-risk can be well bounded by the offset Rademacher complexity. According to our proof strategy, we need to define a covering number to bound the offset Rademacher complexity of the function space \(\gG\).

\begin{Definition}[Modified empirical covering number] Given dataset \(\gS_n\), for any \(f \in \gF\), the empirical metric is defined as
\[
\lVert f \rVert_{\gS_n, d} = \left(\frac{1}{n}\sum_{i=1}^{n} |f(Z_i, Y_i)|^d\right)^{1/d},\ 1 \leq d \leq \infty.
\]

Then, a set \(\gF_\delta\) is termed a \(\delta\)-cover of the hypothesis space \(\gF\) with respect to above empirical metric, if for any \(f \in \gF\), 
there exist \(f_\delta \in \gF_\delta\) such that \(\lVert f - f_\delta \rVert_{\gS_n, d} \leq \delta\). Furthermore, 
\[
N_d(\delta,\gF, \gS_n) = \inf \{|\gF_\delta|: \gF_\delta \ is \ a\ \delta\text{-}cover\ of\ \gF\  with\ respect \ to\  empirical\ metric \}
\]
is the empirical covering number of \(\gF\) with respect to empirical metric conditionally on \(\gS_n\). 
\end{Definition}

In \citet{duan2023fast}, there is a similar definition, but it is conditioned on the input of \(\gS_n\), whereas ours is based on \(\gS_n\). Based on the definition of the modified empirical covering number, we arrive at the following conclusion.

\begin{theorem} \label{theorem32}
Under conditions 1 and 2, the offset Rademacher complexity corresponding to function space \(\gG\) satisfies
\[
\begin{aligned}
\gR^{\text{off}}_n(\gG, \varrho) &=\mathbb{E}\sup_{f\in \gF} \left(\frac{1}{n} \sum_{i=1}^{n} \tau_i g(Z_i, f) -\frac{\varrho}{n} \sum_{i=1}^{n} g^2(Z_i, f) \right) \\
& \leq \frac{1 + \log \mathbb{E} (N_\infty (\delta, \gF, \gS_n))}{2\varrho n} + 4\ell (c_Ac_Z + c_B)(1 + 4\varrho\ell (c_Ac_Z + c_B)).
\end{aligned}
\]
\end{theorem}

Combining Theorems \ref{theorem31} and \ref{theorem32}, we can easily conclude that \(\mathbb{E}_{\gS_n}(R_\phi(\widehat{f}_\phi) - R^*_\phi) = \gO(\log \mathbb{E} (N_\infty (\delta, \gF, \gS_n) / n)\), achieving the optimal convergence rate. While we could further refine the expression of \(N_\infty (\delta, \gF, \gS_n)\), it is not crucial for our purpose here. We omit this for brevity, and interested readers can refer to \citet{vershynin2018high} for more details.

\section{Simulation Studies} \label{sec4}
The main objective of this section is to validate the effectiveness of the proposed method through simulation studies, while examining the impact of the kernel used in the \(D_K\)-distance on classification results. We primarily compare the proposed method with the adaptive method proposed by \citet{wan2023discriminant}, as well as various point-valued time series classification methods. Among these, the point-valued time series classification  methods operate on various representative points within intervals.

\subsection{Simulation design} \label{sec41}
In the simulation studies of classification analysis for univariate and multivariate interval-valued time series, we considered three fundamental univariate interval-valued time series data generation processes (DGPs) \citep{sun2022model, wan2023discriminant}.  For each DGP, we initially generate bivariate point-valued time series in terms of center and range and then reconstruct them into interval-valued time series.  In all DGPs considered, we assume that the residuals of center and range at different time points are independently and identically distributed to a bivariate normal distribution, i.e., \(\epsilon_t = (\epsilon^c_t, \epsilon^r_t)^\top \sim \gN(0, \Sigma), t = 1,2,\cdots\), where \(\Sigma =\left(\begin{array}{cc}
	1 & \rho/2\\
	\rho/ 2 & 1/4
\end{array}\right)\), and \(\rho\) represents the correlation coefficient between the two residuals. We examined five different \(\rho\) values, namely -0.9, -0.5, 0, 0.3, and 0.7. Below are three specific DGPs that we have considered. 

The forms of the three DGPs are as follows:
\[
\begin{aligned}
\text{DGP1:}\  & X_t = (X^c_t, X^r_t)^\top = \sum_{l=1}^{\infty} \pi_l \Phi Z_{t,l} + \epsilon_{t}, \\
\text{DGP2:}\  & X_t = (X^c_t, X^r_t)^\top =\Phi  X_{t-1} + \epsilon_{t} - \Gamma \epsilon_{t-1}, \\
\text{DGP3:}\  & X_t = (X^c_t, X^r_t)^\top =\epsilon_t - \Gamma \epsilon_{t-1}, \\
\end{aligned}
\]
where \(Z_{t,1}= (1, 1)^\top\), \(Z_{t, 2}, Z_{t, 3}, \cdots\) are independently and identically distributed to \(\mathcal{N}(0, \Sigma)\), \(\Phi = \left(\begin{array}{cc}
	0.2 & -0.1\\ 
	0.1 & 0.2
\end{array}\right)\), \(\pi_l = l^{-2} / \sqrt{3}\), and \(\Gamma=\left(\begin{array}{cc}
-0.6 & 0.3\\
0.3 & 0.6
\end{array}\right)\). Then, we will generate simulated data for univariate and multivariate interval-valued time series based on these three fundamental DGPs.

\subsection{Univariate interval-valued time series classification}\label{42}
We consider the following way to generate labeled univariate interval-valued time series datasets. We use one DGP to generate a dataset, where different correlation coefficients correspond to different classes within the dataset, with 500 samples generated for each class. Considering three fundamental DGPs and five different correlation coefficients, we can obtain three datasets, each comprising 5 classes with a total of 2500 samples. Across all datasets, we set the length of interval-valued time series to \(T=150\). For each dataset, we randomly select 80\% as the training set and allocate 20\% as the test set. Additionally, in the imaging process, we first need to determine the trajectory length and time delay, as well as the threshold. For the sake of simplicity, we use default values for the first two parameters from the \texttt{pyts} package and uniformly set the threshold to \(\pi/18\).

Our main focus of comparison lies in the method proposed by \citet{wan2023discriminant}, which essentially uses a convex combination of interval upper and lower bounds as a representation of the interval for subsequent classification, with the convex combination coefficients treated as learnable parameters. To facilitate parameter optimization using stochastic gradient descent, \citet{wan2023discriminant} approximates the Heaviside step function \(H(x)\) with \((1 + \tanh(\nu x)) / 2\). Clearly, \(\nu\) serves as an indicator of the approximation level: the larger \(\nu\), the better the approximation of \((1 + \tanh(\nu x)) / 2\) to \(H(x)\).  Here, we consider five different \(\nu\) values: 1, 5, 10, 15, and 20, to evaluate the impact of varying approximation levels on classification effectiveness. Across all experiments, we use test accuracy as the evaluation metric.

\begin{figure}[H]
\centering
\begin{minipage}[t]{0.19\textwidth}
\centering
\includegraphics[scale=0.52]{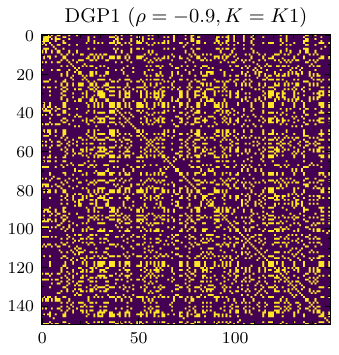}
\end{minipage}
\begin{minipage}[t]{0.19\textwidth}
\centering
\includegraphics[scale=0.52]{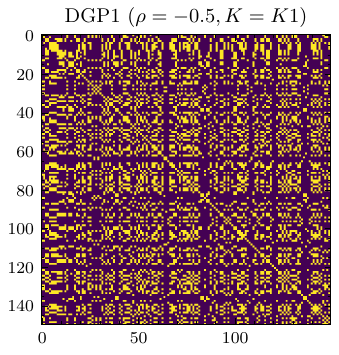}
\end{minipage}
\begin{minipage}[t]{0.19\textwidth}
\centering
\includegraphics[scale=0.52]{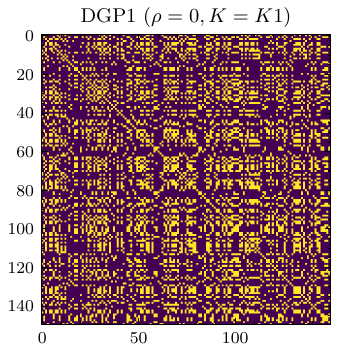}
\end{minipage}
\begin{minipage}[t]{0.19\textwidth}
\centering
\includegraphics[scale=0.52]{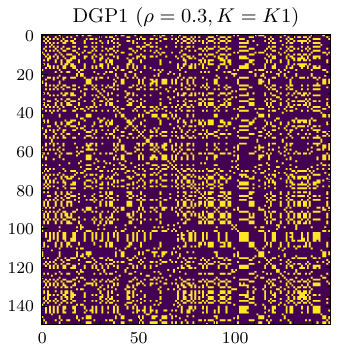}
\end{minipage}
\begin{minipage}[t]{0.19\textwidth}
\centering
\includegraphics[scale=0.52]{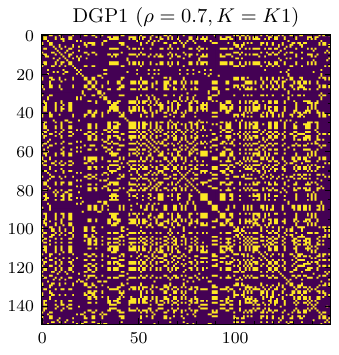}
\end{minipage}

\begin{minipage}[t]{0.19\textwidth}
\centering
\includegraphics[scale=0.52]{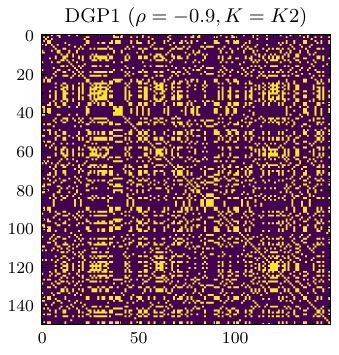}
\end{minipage}
\begin{minipage}[t]{0.19\textwidth}
\centering
\includegraphics[scale=0.52]{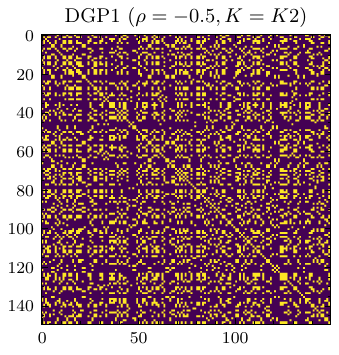}
\end{minipage}
\begin{minipage}[t]{0.19\textwidth}
\centering
\includegraphics[scale=0.52]{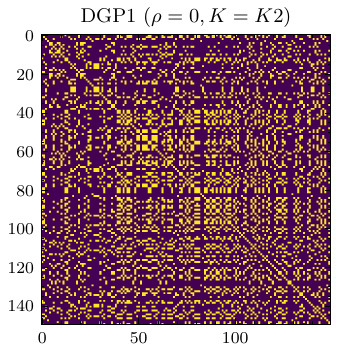}
\end{minipage}
\begin{minipage}[t]{0.19\textwidth}
\centering
\includegraphics[scale=0.52]{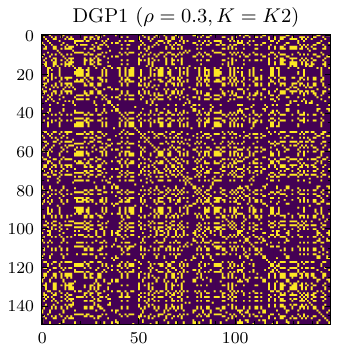}
\end{minipage}
\begin{minipage}[t]{0.19\textwidth}
\centering
\includegraphics[scale=0.52]{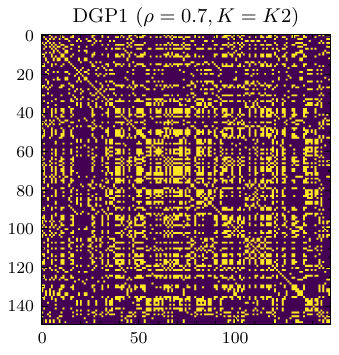}
\end{minipage}

\begin{minipage}[t]{0.19\textwidth}
\centering
\includegraphics[scale=0.52]{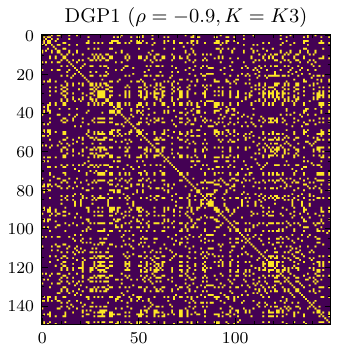}
\end{minipage}
\begin{minipage}[t]{0.19\textwidth}
\centering
\includegraphics[scale=0.52]{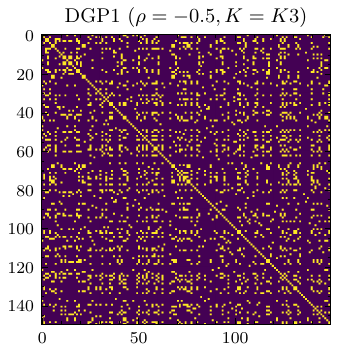}
\end{minipage}
\begin{minipage}[t]{0.19\textwidth}
\centering
\includegraphics[scale=0.52]{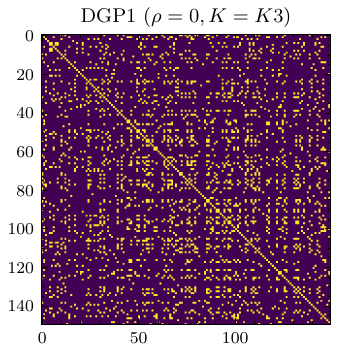}
\end{minipage}
\begin{minipage}[t]{0.19\textwidth}
\centering
\includegraphics[scale=0.52]{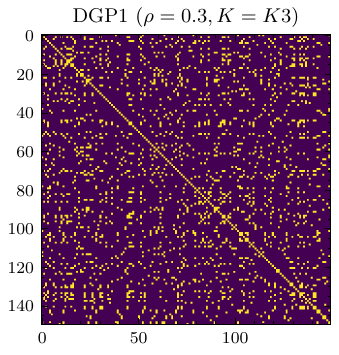}
\end{minipage}
\begin{minipage}[t]{0.19\textwidth}
\centering
\includegraphics[scale=0.52]{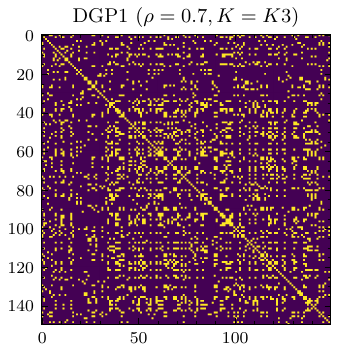}
\end{minipage}
\caption{Imaging results for DGP1 based on IRP under different correlation coefficients and kernel functions. Each row represents the images obtained with different correlation coefficients under the same kernel function. Similarly, each column displays the images for the same correlation coefficient using different kernel functions.}\label{DGP1iaageIRP}
\end{figure}

After obtaining labeled datasets, we utilize the IRP defined in Section \ref{sec22} to convert these input univariate interval-valued time series into images, which are then used for subsequent classification. Image classification involves analyzing the characteristics of the images and selecting appropriate deep learning models. Figure \ref{DGP1iaageIRP} presents the imaging results of DGP1 based on IRP. Clearly, images obtained under the same kernel function but from different correlation coefficients (i.e., different classes) exhibit significant similarities.  This suggests that convolutional neural networks that perform well on typical datasets, such as ResNet \citep{he2016deep}, struggle to achieve good results on these imaging datasets with high intra-class similarity. In this paper, we employ a fine-grained network for classifying imaging data. The network structure we select is WS-DAN \citep{DBLP:journals/corr/abs-1901-09891}, which is also the architecture used by \citet{wan2023discriminant}. 

A prerequisite for classification interval-valued time series using IRP is to specify the kernel function \(K(u, v)\) within the \(D_K\)-distance. We select the following five kernel functions to evaluate their impact on classification performance, 
\[
K_1 = \left[
\begin{array}{cc}
	1/4 & -1/4\\
	-1/4 & 1/4
\end{array}
\right], \ K_2 = \left[
\begin{array}{cc}
	1 & 1\\
	1 & 1
\end{array}
\right], \ K_3 = \left[
\begin{array}{cc}
	1/2 & 1/4\\
	1/4 & 1/2
\end{array}
\right], \ K_4 = \left[
\begin{array}{cc}
	1 & 0 \\
	0 & 1
\end{array}
\right], \ K_5 = \left[
\begin{array}{cc}
	2 & 1\\
	1 & 1
\end{array}
\right].
\]

Clearly, the first four kernel functions \(K\) correspond to the four scenarios discussed in Section \ref{sec2}, utilizing only partial information from the intervals. The final kernel function satisfies the condition (\ref{kernelcondition}), indicating from the discussion of \citet{han2012autoregressive} that this kernel function utilizes all information from the intervals. The first three kernel functions are also discussed in \citet{han2012autoregressive}. It is important to note that in practical application, we do not necessarily require the kernel functions to be positive definite.

The following Table \ref{unRPIRP} presents the classification results based on the method proposed by \citet{wan2023discriminant} and the method introduced in the paper.
\begin{table}[H]
\setlength{\abovecaptionskip}{0pt}%
\setlength{\belowcaptionskip}{3pt}%
\centering
\caption{Classification accuracy based on WS-DAN using RP and IRP} \label{unRPIRP}
\setlength{\tabcolsep}{1mm}{
\begin{tabular}{cccccc|ccccc}
\hline
\multirow{2}{*}{\tabincell{c}{DGP}} & \multicolumn{5}{c|}{\(\nu\)(RP) }  & \multicolumn{5}{c}{kernel (IRP)}\\
\cline{2-11}
&  	\(\nu = 1\) & 	\(\nu = 5\) & 	\(\nu = 10\) & 	\(\nu = 15\) &	\(\nu = 20\) & \(K = K_1\) & \(K = K_2\) & \(K = K_3\) & \(K = K_4\) & \(K = K_5\)\\ 
\hline
DGP1 & 91.67&81.50&88.98&\textbf{95.83}&91.67&85.83&87.08&96.35 & 99.70 & \textbf{99.80} \\
DGP2 & \textbf{85.42}&66.67&75.00&72.92&68.32&87.35&93.33&96.67 & \textbf{99.90} & 99.85 \\
DGP3 &80.56&80.83&83.33&\textbf{85.42}&76.28&83.82&90.62&95.56 & 99.80 & \textbf{99.90}\\	
\hline
\end{tabular}
}
\end{table}

From Table \ref{unRPIRP}, it is noticeable that the classification performance of the method proposed by \citet{wan2023discriminant} does not exhibit a systematic variation with changes in the approximation levels. However, distinct approximation levels show significant differences in classification performance. For instance, in DGP1, achieving the best accuracy at \(\nu = 15\) with 95.83\%, while at \(\nu=5\), it drops to a mere 81.50\%. In the methods proposed in the paper, kernels \(K_4\) and \(K_5\) achieve optimal performance across all three DGPs, with accuracies of 99.80\%, 99.90\%, and 99.90\%, respectively. The first three kernels only utilize partial information from the intervals, exhibiting similar performance across the three DGPs, all considerably lower than the latter two kernels. Comparing these two approaches reveals that, in terms of optimal performance across the three DGPs, methods based on RP consistently underperform those based on IRP. This suggests that utilizing only partial information from the intervals makes it challenging to achieve optimal classification for interval-valued time series.

In addition to the aforementioned results, we also compared our proposed method with various point-valued time series classification methods, such as statistical learning methods and deep learning methods. The premise of using point-valued time series classification methods is to represent intervals using points. In this paper, we selected four representative points—namely, the center (\(c\)), range (\(r\)), upper bound (\(u\)), and lower bound (\(l\))—to represent intervals and utilized them for classification.

We are comparing the five categories of point-valued time series classification methods listed in Section \ref{sec1}. Among the distance-based methods, we include KNeighbors (KN), a modified version of the \(k\)-nearest neighbors classifier tailored for time series data, alongside ShapeDTW (SD) \citep{2017shapeDTW}. Our selection of dictionary-based methods encompasses IndividualBOSS (IB) and IndividualTDE (ITDE). Considering computational efficiency, our choice for the shapelet-based method focuses on MrSQM \citep{Nguyen2021MrSQMFT}. Within the ensemble learning category, we consider Bagging, ComposableTimeSeriesForest (CTF) \citep{DENG2013142}, and WeightedEnsemble (WE). The ongoing advancements in deep learning have led to a myriad of models and techniques for time series classification. In this study, we meticulously selected 16 representative methods from the cutting-edge deep learning-based time series analysis framework \texttt{tsai} for comparative analysis. References and source code for these methods are accessible at \url{https://timeseriesai.github.io/tsai/}. Our diverse selection encompasses ensemble learning and deep learning methods due to their exceptional performance across diverse research scenarios. The methods we have selected for comparison are identical to those chosen in \citet{wan2023discriminant}. The ensuing table presents the classification accuracy achieved by these five method categories, focusing on four representative points within the intervals. 

\begin{table}[H]
\small
\setlength{\abovecaptionskip}{0pt}%
\setlength{\belowcaptionskip}{3pt}%
\centering
\caption{Classification accuracy based on point-value time series methods.} \label{unpointmeth}
\setlength{\tabcolsep}{1mm}{ 
\begin{tabular}{l|cccc|cccc|cccc}
\hline
\multirow{2}{*}{Methods} & \multicolumn{4}{c|}{DGP1} & \multicolumn{4}{c|}{DGP2} & \multicolumn{4}{c}{DGP3}\\
&  \(c\)&\(r\)&\(l\)& \(u\) & \(c\)&\(r\)&\(l\)& \(u\) & \(c\)&\(r\)&\(l\)& \(u\)\\
\hline
KN &19.60&21.40&18.40&20.40&20.60&24.40&18.20&22.60&18.20&22.00&19.60&21.00\\
SD &19.00&19.20&18.34&20.40&19.80&25.60&19.20&23.60&21.80&21.20&19.00&21.20\\
IB &23.60&17.60&21.00&30.80&20.00&23.60&18.80&24.20&19.60&26.00&16.00&24.40\\
ITDE& 19.00&20.80&19.60&23.80&19.60&33.20&21.80&27.60&19.00&32.80&21.80&24.80\\
MrSQM &22.00&18.20&21.60&21.20&21.20&50.80&19.80&44.40&25.80&62.40&18.80&48.40\\
Bagging& 22.40&16.80&21.40&18.60&34.00&42.80&20.60&41.80&27.40&38.80&19.60&43.60\\
CFT &21.60&19.00&82.00&\textbf{86.20}&30.80&45.00&54.00&45.20&31.40&60.60&49.40&65.80\\
WE & 20.20&20.60&19.80&17.20&26.40&38.80&19.80&42.40&25.40&35.60&15.80&43.60\\
\hline 
FCN &22.25 & 24.75 & 83.00& 83.25& 51.75 & 65.25& 64.75& 73.75& 46.00& 71.00& 53.00& 75.25\\
FCNPlus & 21.75&23.50&83.00&84.25&52.50&65.00&64.50&73.75&45.75&72.25&55.25&74.75\\
Inception &22.25&23.75&77.25&80.75&47.75&57.25&58.50&70.25&40.50&64.00&49.25&71.25\\
XCoord & 22.75&21.50&83.75&83.50&49.75&67.00&62.50&73.00&47.00&72.75&54.50&76.00\\
MLP &24.00&23.00&44.50&50.75&23.25&31.00&31.50&29.00&22.00&38.25&23.75&40.50\\
RNN& 22.50&19.75&24.00&28.00&22.50&23.25&22.00&23.25&19.75&22.50&23.25&21.25\\
LSTM &  20.25&20.25&46.00&46.25&32.50&29.75&36.25&41.75&24.00&43.00&37.75&67.00\\
GRU & 21.50&20.25&84.25&85.25&30.00&63.00&44.25&56.25&29.25&64.00&25.50&78.25\\
RNN\_FCN & 22.25&23.75&82.50&81.25&51.25&66.00&62.50&73.00&42.00&67.75&51.75&74.75\\
LSTM\_FCN & 22.25&23.75&81.00&78.50&48.50&62.25&59.75&72.50&48.00&68.75&53.75&\textbf{79.25}\\
GRU\_FCN  & 21.50&23.75&82.00&82.75&50.75&65.00&62.25&71.50&41.50&66.00&53.25&71.00\\
MRNN\_FCN &22.75&24.00&84.00&83.25&50.50&62.75&64.50&72.75&45.25&70.75&51.25&74.75\\
MLSTM\_FCN &23.25&23.75&85.75&80.50&50.00&63.75&62.50&72.75&44.00&69.50&52.50&77.25\\
MGRU\_FCN & 22.25&24.00&82.75&84.75&49.75&64.50&64.25&\textbf{76.00}&42.50&69.25&48.75&72.75\\
ResCNN &24.00&24.00&80.00&82.50&48.50&61.25&61.00&71.00&43.50&67.50&53.50&77.00\\
ResNet & 22.25&22.50&77.25&80.50&48.75&61.00&59.00&67.00&39.00&65.50&49.25&68.50\\
\hline
\end{tabular}
}
\end{table}

From Table \ref{unpointmeth}, it is apparent that both non-deep learning and deep learning methods exhibit remarkably poor classification performance. For instance, all three DPGs achieve optimal performance when the upper bound is the representative point, with accuracies of 86.20\%, 76.00\%, and 79.25\% for CFT, MGRU\_FCN, and LSTM\_FCN, respectively. However, this falls far short. In other scenarios where the combination involves center, range, and lower bound, the accuracy rarely exceeds 70\%. In some cases, it even falls below 25\%. For example, the selected methods for the center and range classification in DGP1. This indicates the near failure or inefficacy of these point-valued time series classification methods in these scenarios. The reason behind these outcomes might be the association of different correlation coefficients and categories during the data generation process, leading to considerable similarities in the resulting interval-valued time series data both within and across classes.

Comparing Tables \ref{unRPIRP} and \ref{unpointmeth}, it is evident that the outcomes obtained from point-based time series classification methods significantly inferior to those proposed by \citet{wan2023discriminant} and the approach presented in this paper, with our method achieving optimal performance. This suggests that relying solely on a simplistic single point to represent intervals falls short in encapsulating crucial information. The relatively better performance of \citet{wan2023discriminant} might stem from its adaptive approach in selecting representative points and utilizing RP for extracting vital information. Our proposed method, leveraging the \(D_K\)-distance and selecting appropriate kernels, thoroughly harnesses interval information, resulting in the current optimal performance. 

\subsection{Multivariate interval-valued time series classification} \label{sec43}
Using the DGPs outlined for univariate interval-valued time series in Subsection \ref{sec41}, we establish two distinct multivariate interval-valued time series generation processes. In the first scenario (C1), each DGP represents a class, while interval-valued time series, generated with varying correlation coefficients \(\rho\), are treated as distinct dimensions. Consequently, this configuration forms a multivariate interval-valued time series dataset comprising three classes and five dimensions. In the second scenario (C2), interval-valued time series generated with diverse correlation coefficients \(\rho\) are viewed as distinct classes, while each DGP functions as a dimension. This configuration yields a multivariate interval-valued time series dataset encompassing five classes and three dimensions. For imaging and classification of multivariate interval-valued time series, we employ the same kernel functions detailed in Subsection \ref{sec21}. Table \ref{siJRPIJRP} delineates the classification performance of multivariate interval-valued time series based on JRP and IJRP for the two aforementioned scenarios. 

\begin{table}[H]
\setlength{\abovecaptionskip}{0pt}%
\setlength{\belowcaptionskip}{3pt}%
\centering
\caption{Classification accuracy based on WS-DAN using JRP and IJRP} \label{siJRPIJRP}
\setlength{\tabcolsep}{1mm}{
\begin{tabular}{cccccc|ccccc}
\hline
\multirow{2}{*}{\tabincell{c}{DGP}} & \multicolumn{5}{c|}{\(\nu\)(JRP) }  & \multicolumn{5}{c}{kernel (IJRP)}\\
\cline{2-11}
&  	\(\nu = 1\) & 	\(\nu = 5\) & 	\(\nu = 10\) & 	\(\nu = 15\) &	\(\nu = 20\) & \(K = K_1\) & \(K = K_2\) & \(K = K_3\) & \(K = K_4\) & \(K = K_5\)\\ 
\hline
C1 & 88.33&89.44&95.67&\textbf{96.43}&88.34&98.33 &94.55&92.48 & \textbf{98.67}& 96.44 \\
C2 &86.72&89.58&\textbf{97.22}&92.36&95.83& 83.55 &73.09&69.58 & 89.08 &\textbf{88.67}\\
\hline
\end{tabular}
}
\end{table}

In this section, we also employed four representative points to represent intervals separately, utilizing five categories of multivariate point-valued time series methods for classification. The corresponding outcomes are presented in Table \ref{simumuipoint} below.

\begin{table}[H]
\setlength{\abovecaptionskip}{0pt}%
\setlength{\belowcaptionskip}{3pt}%
\centering
\caption{Classification accuracy based on point-value time series methods.} \label{simumuipoint}
\setlength{\tabcolsep}{3mm}{ 
\begin{tabular}{l|cccc|cccc}
\hline
\multirow{2}{*}{Methods} & \multicolumn{4}{c|}{C1} & \multicolumn{4}{c}{C2}\\
&  \(c\)&\(r\)&\(l\)& \(u\) & \(c\)&\(r\)&\(l\)& \(u\) \\
\hline
KN& 37.33&38.67&37.33&37.33&17.00&20.80&18.20&20.40\\
ITDE & 71.00  &70.33&65.67&73.00&18.60&38.20&18.60&30.20\\
Bagging & 82.33&75.67&76.67&68.67&23.20&37.40&21.60&35.80\\
WE& 77.00&69.33&73.67&71.33&21.00&34.80&19.60&41.60\\

\hline 
FCN & 98.75& 99.58&98.75&\textbf{100.0}&51.50&77.00& 88.50&\textbf{90.75} \\
Inception & 99.17&99.17& 99.58 &98.75&52.00&75.50&85.25&86.75\\
XCoord & 97.50&98.33&96.25&\textbf{100.0}&54.50& 77.50 &84.50&89.25\\
MLP & 37.92&69.17&50.83&66.67&23.75&26.00&23.50&33.75\\
RNN& 66.67&74.17&48.33&65.83&19.75&22.50&24.25&23.25\\
LSTM & 75.42&86.67&63.33&96.25&31.25&43.00&60.50&68.75\\
GRU & 88.75&96.67&74.58&93.75&36.25&75.25&87.25&89.25\\
RNN\_FCN & 97.50&98.75&96.67&98.33&52.25&70.50&77.75&80.50\\
LSTM\_FCN & 94.58&99.58&94.17&99.58&50.00&72.00&80.00&82.25\\
GRU\_FCN  & 95.83&95.83&92.50&99.58&50.50&68.50&80.00&79.00\\
MRNN\_FCN &98.75&97.50&97.50&98.33& 55.25 &74.75&83.75&84.25\\
MLSTM\_FCN &91.67&96.67&97.50&98.33&54.50&76.00&84.00&84.75\\
MGRU\_FCN &98.75&96.67&93.33&98.75&50.00&77.00&83.25&83.50\\
ResCNN & 97.08&95.00&95.83&99.58&46.75&73.00&84.75&87.75\\
ResNet & 98.33&95.00&95.83&94.58&48.00&71.50&82.00&87.75\\
\hline
\end{tabular}
}
\end{table}

In Table \ref{siJRPIJRP}, similar to the univariate scenario, the classification performance does not exhibit a consistent pattern with the change in approximation level. For instance, higher approximation levels correspond to better outcomes. For C1 and C2, the optimal performance is achieved at \(\nu=15\) and \(\nu=10\), respectively, reaching 96.43\% and 97.22\%. The proposed method in the paper demonstrates excellent classification performance in C1; particularly, with kernel \(K_4\), it achieves the best result of 98.67\%, surpassing the method proposed by \citet{wan2023discriminant}. In the case of C2, the best performance of the proposed method is attained with kernel \(K_4\). Overall, the latter two kernels achieve optimal results by incorporating more interval information. 

Surprisingly, both deep learning and non-deep learning methods demonstrate superior classification performance in multivariate scenarios compared to univariate scenarios, displaying significant enhancements. For instance, in C1, many non-deep learning methods achieve accuracies above 70\% in numerous instances, while several deep learning methods reach accuracies exceeding 95\%. However, the classification performance of these methods in C2 is notably weaker than in C1, which can be gleaned from the generation processes of C1 and C2. Firstly, C1 comprises only three classes, whereas C2 consists of five classes. Secondly, each class in C1 represents distinct univariate interval-valued time series DGP, resulting in lower similarity between classes and reducing the difficulty of classification. Conversely, each class in C2 encompasses three fundamental univariate DGPs, leading to higher similarity between classes. Comparing Tables \ref{siJRPIJRP} and \ref{simumuipoint}, it is apparent that the majority of classification methods based on multivariate point-valued time series did not surpass the optimal performance achieved by the method proposed by \citet{wan2023discriminant} and the method proposed in this paper. Additionally, there is significant variability in performance based on point-based classification methods. This indicates substantial performance disparities among different methods using the same point-based representation. For instance, in C1, KN achieves a classification accuracy of 37.33\% for the upper bound, while FCN and XCoord achieve 100.0\%. Contrasting these methods' classification effectiveness in C2, it is evident that the method proposed in this paper exhibits more advantages in complex scenarios.

\section{Real Data Examples}\label{sec5}
In addition to simulated studies, we also applied our proposed method to real-world naturally occurring interval-valued time series data. Typically, interval-valued data arise from two primary sources: uncertain observations and information aggregation. While there exist studies focusing on classification analysis of interval-valued data, they often operate on data derived through fuzzification of point-valued data, rather than authentic interval-valued data \citep{Utkin2016, Wang2017}. This results in experimental outcomes being associated with the fuzzification method, while the effectiveness of the method on real interval-valued data remains unknown.

The dataset we utilized is sourced from the website \url{https://rp5.ru}. This website contains weather data from 240 countries and regions worldwide. The latest data is added to the site's database at intervals of every three hours, eight times a day. Additionally, the website provides a seven-day weather forecast updated twice daily. In the website's database, there are a total of 28 distinct weather indicators. For this paper, we select six of these indicators for the subsequent classification analysis. Specifically, these selected indicators include the atmospheric temperature at a height of two meters above ground level (measured in degrees Celsius), atmospheric pressure at the weather station level (measured in mmHg), atmospheric pressure at mean sea level (measured in mmHg), relative humidity at a height of two meters above ground level (measured in \%), average wind speed at a height of 10-12 meters above ground level during the ten minutes preceding observation (measured in m/s), and dew point temperature at a height of two meters above ground level (measured in degrees Celsius). We construct intervals by taking the maximum and minimum values of these selected indicators per day as the upper and lower bounds, respectively. Subsequently, we form interval-valued time series with a duration of thirty days for each sequence. 

\subsection{Univariate interval-valued time series classification} \label{sec51}
In the univariate classification scenario, we exclusively choose temperature as the classification variable. We analyze four cities located in various geographical regions across mainland China, each city representing a distinct class. These four cities are Beijing (B) and Harbin (H) in the northern region, Taiyuan (T) in the central region, and Sanya (S) in the southern region. We establish three combinations, H-S, S-T-B, and H-S-T-B, for classification, corresponding to binary and multiclass problems. The selection of smooth parameters and kernel functions, along with the univariate point-valued time series classification methods, aligns with those outlined in Section \ref{42}. The results of these methods are presented in Tables \ref{realRPIRP} and \ref{realunpoint} below.

\begin{table}[H]
\setlength{\abovecaptionskip}{0pt}%
\setlength{\belowcaptionskip}{3pt}%
\centering
\caption{Classification accuracy based on WS-DAN using RP and IRP} \label{realRPIRP}
\setlength{\tabcolsep}{1mm}{
\begin{tabular}{cccccc|ccccc}
\hline
\multirow{2}{*}{\tabincell{c}{DGP}} & \multicolumn{5}{c|}{\(\nu\)(RP) }  & \multicolumn{5}{c}{kernel (IRP)}\\
\cline{2-11}
&  	\(\nu = 1\) & 	\(\nu = 5\) & 	\(\nu = 10\) & 	\(\nu = 15\) &	\(\nu = 20\) & \(K = K_1\) & \(K = K_2\) & \(K = K_3\) & \(K = K_4\) & \(K = K_5\)\\
\hline
H-S & 83.33&91.66&\textbf{92.22}&91.66&91.25&\textbf{100.0}&\textbf{100.0}&\textbf{100.0} &\textbf{100.0} & 99.98\\
S-T-B & 85.52&\textbf{95.83}&94.44&91.67&83.33&86.67&91.67& 95.00  & \textbf{98.26} & 93.02 \\
H-S-T-B	& 70.00&66.67&71.21&\textbf{83.33}&75.00& 88.85 &81.62&85.95 & \textbf{91.44}& 90.24 \\	
\hline
\end{tabular}
}
\end{table}

\begin{table}[H]
\small
\setlength{\abovecaptionskip}{0pt}%
\setlength{\belowcaptionskip}{3pt}%
\centering
\caption{Classification accuracy based on point-value time series methods.} \label{realunpoint}
\setlength{\tabcolsep}{1mm}{ 
\begin{tabular}{l|cccc|cccc|cccc}
\hline
\multirow{2}{*}{Methods} & \multicolumn{4}{c|}{H-S} & \multicolumn{4}{c|}{S-T-B} & \multicolumn{4}{c}{H-S-T-B}\\
&  \(c\)&\(r\)&\(l\)& \(u\) & \(c\)&\(r\)&\(l\)& \(u\) & \(c\)&\(r\)&\(l\)& \(u\)\\
\hline
KN &46.03&55.56&61.90&53.97&25.56&35.34&32.33&27.82&20.53&27.15&22.52&19.87\\
SD&55.56&58.73&60.32&69.84&33.83&48.12&38.35&24.81&23.84&34.43&25.82&21.19\\
IB&61.90&66.67&65.08&68.25&38.35&35.34&39.85&38.35&26.49&33.11&29.14&34.44\\
ITDE&65.08&69.84&61.90 &63.49&37.59&33.08&36.84&36.84&25.83&33.11&28.48&29.14\\
MrSQM&82.54&65.08&74.60&76.19&38.35&43.61&50.38&48.87&43.71&31.13&39.07&36.42\\
Bagging&76.19&80.95&74.60&80.95&39.10&45.86&51.88&50.38&36.42&39.74&42.38&42.38\\
CFT&76.19&87.30&84.13&80.95&41.35&54.14&57.14&46.62&39.07&49.01&50.99&39.07\\
WE&79.37&74.60&88.89&74.60&45.11&40.60&45.11&50.38&35.76&39.07&44.37&38.41\\

\hline 
FCN & \textbf{92.00} & 84.00& 90.00& \textbf{92.00}& 55.66& 59.43& 60.37& 51.89& 52.89& 52.07& 57.02& 55.37\\
FCNPlus & 90.00&86.00&90.00&\textbf{92.00}&55.66&58.49&61.32&59.49&50.41&52.07&57.85&55.37\\ 
Inception & 88.00&86.00&84.00&\textbf{92.00}&59.43&48.11&62.26&51.88&51.24&47.11&51.24&47.11\\
XCoord & 86.00&80.00&86.00&90.00&56.60&53.77&55.66&54.72&51.24&53.72&56.20&55.37\\
MLP & 86.00&76.00&84.00&86.00&56.60&43.39&52.83&45.28&47.10&41.32&46.28&40.50\\
RNN& 84.00&76.00&86.00&86.00&43.40&42.45&44.34&43.40&42.15&41.32&42.98&41.32\\
LSTM & 86.00&76.00&86.00&86.00&41.51&47.17&41.51&43.40&40.50&42.98&38.84&42.98\\
GRU & 86.00&76.00&84.00&86.00&44.34&44.34&44.34&40.57&41.32&42.15&43.80&44.63\\
RNN\_FCN & 90.00&82.00&90.00&\textbf{92.00}&58.49&55.66&59.43&54.72&52.89&52.89&55.37&52.89\\
LSTM\_FCN & \textbf{92.00}&82.00&\textbf{92.00}&\textbf{92.00}&55.66&55.66&57.55&56.60&47.93&52.07&53.72&55.37\\
GRU\_FCN &90.00&84.00&90.00&\textbf{92.00}&54.72&55.66&60.38&56.60&52.07&50.41&55.37&52.07\\
MRNN\_FCN &90.00&84.00&86.00&\textbf{92.00}&55.66&56.60&57.55&56.60&49.59&55.37&53.72&51.24\\
MLSTM\_FCN &90.00&84.00&90.00&\textbf{92.00}&58.49&57.55&60.38&54.72&53.72&53.72&55.37&55.37\\
MGRU\_FCN &90.00&84.00&\textbf{92.00}&\textbf{92.00}&56.60&56.60&60.38&59.43&54.55&51.24&54.55&57.02\\
ResCNN &88.00& 90.00 &\textbf{92.00}&90.00&59.43&58.49&65.09&60.38&53.72&51.24&58.68&54.55\\
ResNet &90.00&86.00&90.00&90.00&59.43&56.60&61.32&56.60&53.72&52.90&56.20&47.93\\
\hline
\end{tabular}
}
\end{table}

Based on Tables \ref{realRPIRP} and \ref{realunpoint}, a comparative analysis of these methods yields several findings. Firstly, the method proposed in this paper achieves the most optimal classification performance, followed by the method proposed by \citet{wan2023discriminant}, while point-valued time series classification methods exhibit the poorest results. Secondly, it appears that the proposed method is not particularly sensitive to specific kernels in binary classification. However, in multi-class scenarios, kernels \(K_4\) and \(K_5\) deliver notably superior results, aligning with the conclusions drawn from the simulation studies. Thirdly, as the number of classes increases, the performance of all three methods experiences varying degrees of decline. For example, the proposed method achieves accuracy rates of 100.0\%, 98.26\%, and 91.44\% across three combinations, whereas point-valued time series methods almost become ineffective in multi-class scenarios.

\subsection{Multivariate interval-valued time series classification} \label{sec52}
Here, we apply the proposed method to classify multivariate interval-valued time series. Apart from utilizing all six selected variables, the construction of interval-valued data and the combinations used for classification remain consistent with Section \ref{sec51}. Additionally, our selection of smoothing parameters, kernel functions, and the approach for multivariate point-valued time series classification aligns with Section \ref{sec43}. The results of these methods are displayed in Tables \ref{realJRPIJRP} and \ref{realmupoint} below.

\begin{table}[H]
\setlength{\abovecaptionskip}{0pt}%
\setlength{\belowcaptionskip}{3pt}%
\centering
\caption{Classification accuracy based on WS-DAN using JRP and IJRP} \label{realJRPIJRP}
\setlength{\tabcolsep}{1mm}{
\begin{tabular}{cccccc|ccccc}
\hline
\multirow{2}{*}{\tabincell{c}{DGP}} & \multicolumn{5}{c|}{\(\nu\)(JRP) }  & \multicolumn{5}{c}{kernel (IJRP)}\\
\cline{2-11}
& \(\nu = 1\) &  \(\nu = 5\) & 	\(\nu = 10\) & 	\(\nu = 15\) &	\(\nu = 20\) & \(K = K_1\) & \(K = K_2\) & \(K = K_3\) & \(K = K_4\) & \(K = K_5\)\\ 
\hline
H-S & 95.83&95.83&93.33&89.58&\textbf{100.00}&69.71&70.47& 71.19 & \textbf{95.54} & 91.72\\
S-T-B & \textbf{98.96}&97.22&97.22&91.67&93.52&72.00& 73.00 &72.62 & 91.89  & \textbf{95.47}\\
H-S-T-B	& 87.50 &84.72&\textbf{93.75}&87.50&91.67&73.62&73.89& 73.90 & 89.74 & \textbf{95.07}\\	
\hline
\end{tabular}
}
\end{table}

\begin{table}[H]
\scriptsize
\setlength{\abovecaptionskip}{0pt}%
\setlength{\belowcaptionskip}{3pt}%
\centering
\caption{Classification accuracy based on point-value time series methods.} \label{realmupoint}
\setlength{\tabcolsep}{0.2mm}{ 
\begin{tabular}{l|cccc|cccc|cccc|cccc}
\hline
\multirow{2}{*}{Methods} & \multicolumn{4}{c|}{H-S} & \multicolumn{4}{c|}{S-T-B} & \multicolumn{4}{c|}{H-S-T-B} & \multicolumn{4}{c}{Fine-grained}\\
&  \(c\)&\(r\)&\(l\)& \(u\) & \(c\)&\(r\)&\(l\)& \(u\) & \(c\)&\(r\)&\(l\)& \(u\) & \(c\)&\(r\)&\(l\)& \(u\) \\
\hline
KN &92.06&90.48&88.89&92.06&\textbf{100.0}0&66.17&\textbf{100.00}&98.50&96.03&67.55&96.69&93.38&77.78&94.44&77.78&90.74\\
ITDE &76.19&69.84&80.95&82.54&64.66&42.86&59.40&72.18&52.98&46.36&56.29&58.28&51.85&66.67&59.26&50.00\\
Bagging &73.02&79.37&71.43&73.02&31.58&36.84&36.09&33.83&29.80 &37.75&36.42&47.68&83.33&83.33&83.33&83.33\\
WE & 93.65&74.60&87.30&90.48&58.65&48.12& 58.65&56.39&61.59&45.70&58.28&54.97&83.33&83.33&83.33&83.33\\
\hline 
FCN & \textbf{100.0}&\textbf{100.0}&\textbf{100.0}&\textbf{100.0}&\textbf{100.0}&\textbf{100.0}&\textbf{100.0}&\textbf{100.0}&\textbf{99.17}&\textbf{99.17}&96.69&\textbf{99.17}&\textbf{97.62}&95.24&92.86&\textbf{97.62} \\
Inception & 98.00&96.00&98.00&96.00&98.11&94.34&97.17&91.50&97.52&\textbf{99.17}&\textbf{99.17}&98.35&92.86&95.24&88.10&95.24\\
XCoord & 94.00&92.00&94.00&96.00&92.45&95.28&98.11&98.11&98.35&97.52&98.35&97.52&90.48&90.48&90.48&92.86\\
MLP & 88.00&94.00&86.00&98.00&97.17&93.40&84.90&93.40&94.21&96.69&95.04&96.69&85.71&\textbf{97.62}&85.71&95.24\\
RNN& 92.00&94.00&92.00&90.00&89.62&93.40&90.57&95.28&86.78&94.22&92.56&89.27&88.10&95.24&88.10&95.24\\
LSTM & 88.00&94.00&92.00&94.00&91.51&99.06&\textbf{100.0}&98.11&95.04&95.04&95.87&95.87&88.10&97.62&88.10&\textbf{97.62}\\
GRU & 86.00&98.00&92.00&96.00&97.17&98.11&98.11&99.06&94.22&95.04&95.04&95.04&88.10&95.24&88.10&95.24\\
RNN\_FCN & 94.00&96.00&94.00&98.00&99.06&98.11&99.06&98.11&96.70&98.35&98.35&99.17&97.62&97.62&92.86&95.24\\
LSTM\_FCN & 96.00&98.00&98.00&98.00&99.06&99.06&99.06&98.11&96.69&\textbf{99.17}&\textbf{99.17}&\textbf{99.17}&95.24&97.62&92.86&92.86\\
GRU\_FCN  &98.00&98.00&98.00&98.00&99.06&98.11&91.51&85.85&91.74&98.35&97.52&98.35&88.10&88.10&90.48&92.86\\
MRNN\_FCN &94.00&96.00&96.00&98.00&96.23&98.11&87.74&99.06&98.35&98.35&98.35&98.35&88.10&97.62&92.86&\textbf{97.62}\\
MLSTM\_FCN &94.00&96.00&96.00&98.00&92.45&98.11&96.22&99.06&98.35&98.35&98.35&97.52&90.48&95.24& 95.24&95.24\\
MGRU\_FCN & 92.00&98.00&96.00&98.00&99.06&98.11&98.11&99.06&98.35&\textbf{99.17}&97.52&\textbf{99.17}&95.24&97.62&92.86&\textbf{97.62}\\
ResCNN &96.00&98.00&96.00&98.00&98.11&98.11&94.34&99.06&95.04&98.35&98.35&98.35&95.24&95.24&90.48&92.86\\
ResNet &96.00&96.00&94.00&96.00&98.11&98.11&98.11&99.06&95.87&95.87&96.69&\textbf{99.17}&95.24&95.24&90.48&95.24\\
\hline
\end{tabular}
}
\end{table}

Comparing Tables \ref{realJRPIJRP} and \ref{realmupoint}, it can be observed that there is little difference in the classification performance among the method proposed in this paper, the method proposed by \citet{wan2023discriminant}, and the multivariate point-value time series methods. In some scenarios, the multivariate point-value time series methods perform even better. For instance, FCN achieves a perfect accuracy of 100\% on four representative points in combinations H-S and S-T-B, and it reaches 99.17\% accuracy in combination H-S-T-B. However, non-deep learning methods show poorer performance, with accuracy rates in many scenarios not even exceeding 50\%. Similar to previous observations, our proposed method also demonstrates optimal performance on kernels \(K_4\) and \(K_5\), which make full use of interval information.

Comparing these results with the classification outcomes of univariate interval time series in Section \ref{sec51} reveals a significant improvement in their respective classification effects as the number of variables increases. For example, in both univariate and multivariate scenarios, the FCN method achieves optimal accuracy rates of 92.00\% and 100.0\%, respectively, for the combination of H-S. Considering the comprehensive analysis above, to achieve a better classification effect, apart from choosing an appropriate method, providing a sufficient description of the original problem with an adequate number of variables is also necessary.

\subsection{Fine-grained classification} \label{sec53}
Besides the conventional classification analysis mentioned earlier, we proceeded to assess the effectiveness of the proposed method in fine-grained classification analysis. Fine-grained classification refers to substantial similarity both within and between classes in the data. However, the four cities selected above are quite distant, resulting in considerable differences between these classes. For this purpose, we collected six air quality indicators' data from \url{https://rp5.ru/} for Shanghai Baoan Station spanning from January 2, 2005, to April 11, 2023, and Shanghai Hongqiao Station from July 7, 2019, to April 11, 2023. Subsequently, we aggregated this data on a daily basis into multivariate interval-valued time series. Similar to Section \ref{sec51}, we consider observations for 30 days as trajectories to construct multivariate interval-valued time series classification data for each region. Additionally, it is notable from the data obtained on the website that the volume of data for Hongqiao district is significantly smaller compared to Baoan district. This exacerbates class imbalance in the context of fine-grained classification, further complicating the issue.
The classification results based on these methods are displayed in the final column of Table \ref{realmupoint} and Table \ref{realfenJRPIJRP} below.

\begin{table}[H]
\setlength{\abovecaptionskip}{0pt}%
\setlength{\belowcaptionskip}{3pt}%
\centering
\caption{Classification accuracy based on WS-DAN using JRP and IJRP} \label{realfenJRPIJRP}
\setlength{\tabcolsep}{1mm}{
\begin{tabular}{ccccc|ccccc}
\hline
\multicolumn{5}{c|}{\(\nu\)(JRP) }  & \multicolumn{5}{c}{kernel (IJRP)}\\
\hline
\(\nu = 1\) & 	\(\nu = 5\) & 	\(\nu = 10\) & 	\(\nu = 15\) &	\(\nu = 20\) & \(K = K_1\) & \(K = K_2\) & \(K = K_3\) &\(K = K_4\) & \(K = K_5\)\\ 
\hline		
91.67 & 83.34 & 87.50 & 91.67&  \textbf{95.83} & 86.11 & 87.50 & 87.50  & \textbf{89.62} & 89.15\\
\hline
\end{tabular}
}
\end{table}

In fine-grained classification, we employed a set of six weather indicators, essentially constituting a multivariate interval-valued time series classification task. Upon comparing the experimental outcomes derived from these methods with those obtained in Section \ref{sec52}, it is noticeable that the performance of these methods has shown a certain degree of decline in fine-grained classification, although their accuracy remains notably high. For instance, the method proposed in this paper achieves a maximum accuracy of 89.62\%. Moreover, these experimental findings showcase resemblances to those observed in Section \ref{sec52}, suggesting the enduring effectiveness and relative robust of the proposed method within complex	 scenarios.

\section{Conclusions} \label{sec6}
In this paper, we introduced a classification method that treats intervals as a unified entity, suitable for both univariate and multivariate interval-valued time series. At the core of this method lies the use of the \(D_K\)-distance as a measure for distances between intervals. Building upon this concept, we extended point-based time series imaging methods to encompass interval-valued scenarios, aiming to extract crucial information from interval-valued time series for subsequent classification. Both in simulated studies and real-data applications, we observed that the proposed method excels in most scenarios, followed by the method proposed by \citet{wan2023discriminant} and lastly, the point-valued time series classification methods. These outcomes indicate that representing intervals using a single, simplistic point leads to the loss of a significant portion of information contained within the intervals, resulting in the ineffectiveness of classification methods.

We should note that in addition to the \(D_K\)-distance, the Hausdorff metric \citep{munkres2018elements} can also serve as a distance measure between pairwise intervals. Furthermore, besides RP and JRP, there are other point-based time series imaging methods such as Gramian Angular Summation Field, Gramian Angular Difference Field, and Markov Transition Field \citep{wang2015encoding}. These aspects warrant further in-depth research in the future.

\section*{Acknowledgments}
\addcontentsline{toc}{section}{Acknowledgments}
The research work described in this paper was supported by the National Natural Science Foundation of China (Nos. 72071008). 

\newpage

\begin{appendix}		
\section{Proofs for Results}\label{appendix}

{\bf Lemma 3.1.} {
\it Suppose Conditions 1 and 2 hold, then for any \(f_1, f_2 \in \gF\), we have
\[
|g(Z, f_1)| \leq 2\ell (c_Ac_Z + c_B), \ |g(Z, f_1) - g(Z, f_2)| \leq 4\ell (c_Ac_Z + c_B),
\]
where \(\ell\) is the Lipschitz constant of the auxiliary function \(L(\cdot)\). 
}

\begin{proof}
For any \(f_1, f_2 \in \gF\),  we have
\[
\begin{aligned}
|g(Z, f_1) - g(Z, f_2)| & = |\mathbb{E}(\phi(f_1, Z, Y) - \phi(f^*_\phi, Z, Y)|Z) - \mathbb{E}(\phi(f_2, Z, Y) - \phi(f^*_\phi, Z, Y)|Z)|\\
& = |\mathbb{E}(\phi(f_1, Z, Y) - \phi(f_2, Z, Y)|Z)|\\
& = |\mathbb{E}(L(f_1 (Z, Y) - \max_{Y^\prime \neq Y} f_1(Z, Y^\prime)) - L(f_2(Z, Y) - \max_{Y^\prime \neq Y} f_2(Z, Y^\prime))|Z)|\\
& \leq \mathbb{E}(|L(f_1 (Z, Y) - \max_{Y^\prime \neq Y} f_1(Z, Y^\prime)) - L(f_2(Z, Y) - \max_{Y^\prime \neq Y} f_2(Z, Y^\prime))| |Z)\\
& \leq \mathbb{E}(\ell|(f_1 (Z, Y) - \max_{Y^\prime \neq Y} f_1(Z, Y^\prime)) -  (f_2(Z, Y) - \max_{Y^\prime \neq Y} f_2(Z, Y^\prime))| | Z )\\
& \leq \ell \mathbb{E}(|f_1 (Z, Y) - \max_{Y^\prime \neq Y} f_1(Z, Y^\prime)| | Z) + \ell \mathbb{E}(|f_2 (Z, Y) - \max_{Y^\prime \neq Y} f_2(Z, Y^\prime)| | Z)\\
& \leq 2\ell \mathbb{E} \left(\sup_{A, B, Z} |f_1 (Z, Y) - \max_{Y^\prime \neq Y} f_1(Z, Y^\prime)| | Z\right) \\
& \leq 4 \ell \sup_{A, B, Z} |f_1 (Z, Y)| \\
& = 4 \ell\sup_{A, B, Z} |A_Y^\top Z + B_Y | \\
& \leq 4 \ell \sup_{A, B, Z} (|A_Y^\top Z| + |B_Y|)\\
& \leq 4\ell (c_Ac_Z + c_B).
\end{aligned}
\]

Setting \(f_2 = 0\), the first inequality can be immediately obtained. This completes the proof.
\end{proof}

{\bf Theorem 3.1} {\it 
Suppose conditions 1 and 2 hold, the excess \(\phi\)-risk satisfies
\[
\mathbb{E}_{\gS_n}(R_\phi(\widehat{f}_\phi) - R^*_\phi) \leq  4 \mathcal{R}^{\text{off}}_n\left(\gG, \frac{1}{4\ell (c_Ac_Z + c_B)}\right).
\]
}
\begin{proof}

Using the fact that \(\widehat{f}_\phi\) be the empirical \(\phi\)-risk minimizer, we have
\[
\begin{aligned}
\mathbb{E}_{Z, Y}(\phi(\widehat{f}_\phi, Z, Y) - \phi(f^*_\phi, Z, Y))  & = \mathbb{E}_{Z, Y}(\phi(\widehat{f}_\phi, Z, Y) - \phi(f^*_\phi, Z, Y))- \frac{3}{n}\sum_{i=1}^{n} (\phi(\widehat{f}_\phi, Z_i, Y_i) - \phi(f^*_\phi, Z_i, Y_i)) \\
& + \frac{3}{n}\sum_{i=1}^{n} (\phi(\widehat{f}_\phi, Z_i, Y_i) - \phi(f^*_\phi, Z_i, Y_i))\\
& \leq \mathbb{E}_{Z, Y}(\phi(\widehat{f}_\phi, Z, Y) - \phi(f^*_\phi, Z, Y))- \frac{3}{n}\sum_{i=1}^{n} (\phi(\widehat{f}_\phi, Z_i, Y_i) - \phi(f^*_\phi, Z_i, Y_i)) \\
& + \frac{3}{n}\sum_{i=1}^{n} (\phi(f, Z_i, Y_i) - \phi(f^*_\phi, Z_i, Y_i)).\\
\end{aligned} 
\]

The expectation over the dataset \(\gS_n\) on both sides of the above inequality yields
\[
\begin{aligned}
\mathbb{E}_{\gS_n} & \left(\mathbb{E}_{Z, Y}(\phi(\widehat{f}_\phi, Z, Y) - \phi(f^*_\phi, Z, Y))\right)  \leq \mathbb{E}_{\gS_n} \bigg(
\mathbb{E}_{Z, Y}(\phi(\widehat{f}_\phi, Z, Y) - \phi(f^*_\phi, Z, Y)) \\
& - \frac{3}{n}\sum_{i=1}^{n} (\phi(\widehat{f}_\phi, Z_i, Y_i) - \phi(f^*_\phi, Z_i, Y_i))  + \frac{3}{n}\sum_{i=1}^{n} (\phi(f, Z_i, Y_i) - \phi(f^*_\phi, Z_i, Y_i)) \bigg)\\
& \leq \sup_{f\in \gF} \mathbb{E}_{\gS_n} \bigg(
\mathbb{E}_{Z, Y}(\phi(f, Z, Y) - \phi(f^*_\phi, Z, Y))  - \frac{3}{n}\sum_{i=1}^{n} (\phi(f, Z_i, Y_i) - \phi(f^*_\phi, Z_i, Y_i)) \bigg)\\
& + 3\mathbb{E}_{Z, Y}(\phi(f, Z, Y) - \phi(f^*_\phi, Z, Y))\\
& \leq \mathbb{E}_{\mathbb{Z}} \sup_{f\in \gF} \bigg(
\mathbb{E}_{Z}(g(Z, f))  - \frac{3}{n}\sum_{i=1}^{n} (g(Z_i, f)) \bigg),\\
\end{aligned}
\]
where \(\mathbb{Z} = \{Z_i\}^n_{i=1}\), the final inequality holds by utilizing the convexity of the supremum and the fact that \(f^*_\phi\) is the minimizer corresponding to the optimal \(\phi\)-risk.  From Lemma 1, we can derive that
\[
0 \leq g(Z, f) \leq 2\ell (c_Ac_Z + c_B),\  g^2(Z, f) \leq 2\ell (c_Ac_Z + c_B) g(Z, f)
\]
holds for any \(f\in \gF\) and \( Z\in \gZ\). Therefore, we have
\[
\begin{aligned}
& \mathbb{E}_{\mathbb{Z}} \sup_{f\in \gF} \bigg(
\mathbb{E}_{Z}g(Z, f) - \frac{3}{n}\sum_{i=1}^{n} g(Z_i, f) \bigg) \\
& = \mathbb{E}_{\mathbb{Z}}  \sup_{f\in \gF} \bigg(
2\mathbb{E}_{Z}g(Z, f)  - \mathbb{E}_{Z}g(Z, f) - \frac{1}{n}\sum_{i=1}^{n}g(Z_i, f)- \frac{2 }{n}\sum_{i=1}^{n}g(Z_i, f)\bigg) \\
& \leq \mathbb{E}_{\mathbb{Z}}  \sup_{f\in \gF} \bigg(
2\mathbb{E}_{Z}g(Z, f)  - \frac{1}{2\ell (c_Ac_Z + c_B)}\mathbb{E}_{Z}g^2(Z, f) - \frac{2}{n}\sum_{i=1}^{n}g(Z_i, f)- \frac{1}{2n\ell (c_Ac_Z + c_B)}\sum_{i=1}^{n}g^2(Z_i, f)\bigg).
\end{aligned}
\]

Let \(\mathbb{Z}^\prime =\{Z^\prime_i\}^n_{i=1}\) be independent identically distributed copies of \(\mathbb{Z}\), then we have
\[
\begin{aligned}
& \mathbb{E}_{\mathbb{Z}}  \sup_{f\in \gF} \bigg(
2\mathbb{E}_{Z}g(Z, f)  - \frac{1}{2\ell (c_Ac_Z + c_B)}\mathbb{E}_{Z}g^2(Z, f) - \frac{2}{n}\sum_{i=1}^{n}g(Z_i, f)- \frac{1}{2n\ell (c_Ac_Z + c_B)}\sum_{i=1}^{n}g^2(Z_i, f)\bigg)\\
& = \mathbb{E}_{\mathbb{Z}}  \sup_{f\in \gF} \left(
\mathbb{E}_{\mathbb{Z}^\prime}\left(
\frac{2}{n}\sum_{i=1}^{n}g(Z^\prime_i, f)  - \frac{1}{2n\ell (c_Ac_Z + c_B)} \sum_{i=1}^{n} g^2(Z^\prime_i, f)
\right) \right)\\
& - \frac{2}{n}\sum_{i=1}^{n}g(Z_i, f)- \frac{1}{2n\ell (c_Ac_Z + c_B)}\sum_{i=1}^{n}g^2(Z_i, f)\bigg)\\
& \leq \mathbb{E}_{\mathbb{Z}} \mathbb{E}_{\mathbb{Z}^\prime} \sup_{f\in \gF} \left(
\frac{2}{n} \sum_{i=1}^{n}(g(Z^\prime_i, f) - g(Z_i, f)) - \frac{1}{2n\ell (c_Ac_Z + c_B)}\sum_{i=1}^{n} (g^2(Z^\prime_i, f) + g^2(Z_i, f))
\right)\\
& = 2\mathbb{E}_{\mathbb{Z}^\prime} \mathbb{E}_\tau \sup_{f\in \gF} \frac{1}{n} \sum_{i=1}^{n} \left(
\tau_i g(Z^\prime_i, f) - \frac{1}{4\ell (c_Ac_Z + c_B)} g^2(Z^\prime_i, f)\right) \\
& +  2\mathbb{E}_{\mathbb{Z}} \mathbb{E}_\tau \sup_{f\in \gF} \frac{1}{n} \sum_{i=1}^{n} \left(
-\tau_i g(Z_i, f) - \frac{1}{4\ell (c_Ac_Z + c_B)} g^2(Z_i, f)\right) \\
& = 2 \mathcal{R}^{\text{off}}_n\left(\gG, \frac{1}{4\ell (c_Ac_Z + c_B)}\right) +  2 \mathcal{R}^{\text{off}}_n\left(\gG, \frac{1}{4\ell (c_Ac_Z + c_B)}\right)\\
& = 4 \mathcal{R}^{\text{off}}_n\left(\gG, \frac{1}{4\ell (c_Ac_Z + c_B)}\right), 
\end{aligned}
\]
where the first inequality is due to the convexity of the supremum, and the introduction of \(\mathbb{Z}^\prime\) is to utilize symmetrization technique.
\end{proof}

{\bf Theorem 3.2} {\it Under conditions 1 and 2, the offset Rademacher complexity corresponding to function space \(\gG\) satisfies
\[
\begin{aligned}
	\gR^{\text{off}}_n(\gG, \varrho) &=\mathbb{E}\sup_{f\in \gF} \left(\frac{1}{n} \sum_{i=1}^{n} \tau_i g(Z_i, f) -\frac{\varrho}{n} \sum_{i=1}^{n} g^2(Z_i, f) \right) \\
	& \leq \frac{1 + \log \mathbb{E} (N_\infty (\delta, \gF, \gS_n))}{2\varrho n} + 4\ell (c_Ac_Z + c_B)(1 + 4\varrho\ell (c_Ac_Z + c_B)).
\end{aligned}
\]
} 

\begin{proof}
Let \(\gF_\delta\) be a \(\delta\)-cover of \(d\to \infty\) of the hypothesis space \(\gF\). That is, for any \(f \in \gF\), there exists \(f_\delta\) such that \(\lVert f - f_\delta \rVert_{\gS_n, \infty} \leq \delta\). We have
\begin{equation} \label{theorem321}
\begin{aligned}
	\frac{1}{n} \sum_{i=1}^{n} \tau_i g(Z_i, f) &\leq \frac{1}{n} \sum_{i=1}^{n} \tau_i g(Z_i, f_\delta) + \frac{1}{n} \sum_{i=1}^{n} |\tau_i| |g(Z_i, f) -g(Z_i, f_\delta)|\\
	& \leq \frac{1}{n} \sum_{i=1}^{n} \tau_i g(Z_i, f_\delta) + 4\ell (c_Ac_Z + c_B),
\end{aligned}
\end{equation}
where the second inequality is obtained using the second inequality from Lemma 1. Furthermore, we have
\begin{equation}\label{theorem322}
\begin{aligned}
-\frac{1}{n} \sum_{i=1}^{n} g^2(Z_i, f) &= -\frac{1}{n} \sum_{i=1}^{n} g^2(Z_i, f_\delta) + \frac{1}{n} \sum_{i=1}^{n} (g(Z_i, f_\delta) + g(Z_i, f))(g(Z_i, f_\delta) - g(Z_i, f)) \\
& \leq -\frac{1}{n} \sum_{i=1}^{n}  g^2(Z_i, f_\delta)  + 4\ell (c_Ac_Z + c_B) \frac{1}{n} \sum_{i=1}^{n}|g(Z_i, f_\delta) - g(Z_i, f)|\\
& \leq -\frac{1}{n} \sum_{i=1}^{n}  g^2(Z_i, f_\delta)  + 4\ell (c_Ac_Z + c_B) \times 4\ell (c_Ac_Z + c_B) \\
& = -\frac{1}{n} \sum_{i=1}^{n}  g^2(Z_i, f_\delta)  + 16\ell^2 (c_Ac_Z + c_B)^2,  
\end{aligned}
\end{equation}
where the first and second inequalities respectively utilize the first and second inequalities of Lemma 1. Setting \(\tau = \{\tau_i\}^n_{i=1}\). Combining (\ref{theorem321}) and (\ref{theorem322}), we have
\begin{equation}\label{theorem323} 
\begin{aligned}
	\mathbb{E}_\tau\sup_{f\in \gF} \left(\frac{1}{n} \sum_{i=1}^{n} \tau_i g(Z_i, f) -\frac{\varrho}{n} \sum_{i=1}^{n} g^2(Z_i, f) \right) &\leq \mathbb{E}_\tau \sup_{f_\delta\in \gF_\delta}\left(\frac{1}{n} \sum_{i=1}^{n} \tau_i g(Z_i, f_\delta) -\frac{\varrho}{n} \sum_{i=1}^{n} g^2(Z_i, f_\delta)\right)\\
	& + 4\ell (c_Ac_Z + c_B)(1 + 4\varrho\ell (c_Ac_Z + c_B)).
\end{aligned}
\end{equation}

It is clear that \(\{\tau_i g(Z_i, f_\delta)\}^n_{i=1}\) are independent identically distributed random variables given \(\gS_n\). We have
\[
\mathbb{E}_\tau(\tau_i g(Z_i, f_\delta)) = 0, \  - g(Z_i, f_\delta) \leq \tau_i g(Z_i, f_\delta) \leq g(Z_i, f_\delta), i = 1,2,\cdots, n,
\]
then, for any \(f_\delta \in \gF_\delta\) and \(\xi\), using Hoeffding's inequality, we obtain
\begin{equation}\label{theorem324}
\begin{aligned}
	\mathbb{P}_\tau \left(\frac{1}{n} \sum_{i=1}^{n} \tau_i g(Z_i, f_\delta) \geq \xi + \frac{\varrho}{n} \sum_{i=1}^{n} g^2(Z_i, f_\delta)\right) &\leq \exp\left(
	- \frac{(n\xi + \varrho \sum_{i=1}^{n} g^2(Z_i, f_\delta))^2}{2\sum_{i=1}^{n} g^2(Z_i, f_\delta)}
	\right)\\
	& \exp(-2\varrho n\xi),
\end{aligned}
\end{equation}
where the second inequality is based on the fact that \((a+b)^2 / b \geq (a +a)^2/a= 4a\) for \(a > 0\). Next, we estimate the first term on the right-hand side of the inequality (\ref{theorem323}), 
\[
\begin{aligned}
& \mathbb{E}_\tau \sup_{f_\delta\in \gF_\delta}\left(\frac{1}{n} \sum_{i=1}^{n} \tau_i g(Z_i, f_\delta) -\frac{\varrho}{n} \sum_{i=1}^{n} g^2(Z_i, f_\delta)\right)\\
& \leq \int_{0}^{\infty} \mathbb{P}_\tau \left(\sup_{f_\delta\in \gF_\delta}\left(\frac{1}{n} \sum_{i=1}^{n} \tau_i g(Z_i, f_\delta) -\frac{\varrho}{n} \sum_{i=1}^{n} g^2(Z_i, f_\delta) > \xi \right)\right) d \xi \\
& \leq h + \int_{h}^{\infty} \mathbb{P}_\tau \left(\sup_{f_\delta\in \gF_\delta}\left(\frac{1}{n} \sum_{i=1}^{n} \tau_i g(Z_i, f_\delta) -\frac{\varrho}{n} \sum_{i=1}^{n} g^2(Z_i, f_\delta) > \xi \right)\right) d \xi \\
& \leq h + \int_{h}^{\infty} N_\infty (\delta, \gF, \gS_n) \max_{f_\delta\in \gF_\delta}\mathbb{P}_\tau\left(\frac{1}{n} \sum_{i=1}^{n} \tau_i g(Z_i, f_\delta) \geq \xi + \frac{\varrho}{n} \sum_{i=1}^{n} g^2(Z_i, f_\delta)\right) \\
& \leq A + \int_{h}^{\infty} N_\infty (\delta, \gF, \gS_n) \exp(-2\varrho n\xi) d\xi\\
& = A+ \frac{N_\infty (\delta, \gF, \gS_n)}{2\varrho n} \exp(-2\varrho nh),
\end{aligned}
\]
by setting \(A = \frac{\log N_\infty (\delta, \gF, \gS_n)}{2\varrho n}\) leads to
\[
\mathbb{E}_\tau \sup_{f_\delta\in \gF_\delta}\left(\frac{1}{n} \sum_{i=1}^{n} \tau_i g(Z_i, f_\delta) -\frac{\varrho}{n} \sum_{i=1}^{n} g^2(Z_i, f_\delta)\right) \leq \frac{1 + \log N_\infty (\delta, \gF, \gS_n)}{2\varrho n}.
\]

Combining the above inequality with (\ref{theorem322}), we obtain
\[
\mathbb{E}\sup_{f\in \gF} \left(\frac{1}{n} \sum_{i=1}^{n} \tau_i g(Z_i, f) -\frac{\varrho}{n} \sum_{i=1}^{n} g^2(Z_i, f) \right) \leq \frac{1 + \log \mathbb{E} (N_\infty (\delta, \gF, \gS_n))}{2\varrho n} + 4\ell (c_Ac_Z + c_B)(1 + 4\varrho\ell (c_Ac_Z + c_B)).
\]

\end{proof}

\end{appendix}

\newpage 
\bibliography{./bib/reference.bib}

\begin{thebibliography}{}

\bibitem[Arroyo et~al., 2011]{arroyo2011different}
Arroyo, J., Esp{\'\i}nola, R., and Mat{\'e}, C. (2011).
\newblock Different approaches to forecast interval time series: a comparison
  in finance.
\newblock {\em Computational Economics}, 37:169--191.

\bibitem[Arroyo et~al., 2007]{arroyo2007exponential}
Arroyo, J., San~Roque, A.~M., Mat{\'e}, C., and Sarabia, A. (2007).
\newblock Exponential smoothing methods for interval time series.
\newblock In {\em Proceedings of the 1st European Symposium on Time Series
  Prediction}, pages 231--240.

\bibitem[Awasthi et~al., 2022]{awasthi2022multi}
Awasthi, P., Mao, A., Mohri, M., and Zhong, Y. (2022).
\newblock {Multi-Class $H$-Consistency Bounds}.
\newblock {\em Advances in neural information processing systems}, 35:782--795.

\bibitem[Bagnall et~al., 2017]{bagnall2017great}
Bagnall, A., Lines, J., Bostrom, A., Large, J., and Keogh, E. (2017).
\newblock The great time series classification bake off: a review and
  experimental evaluation of recent algorithmic advances.
\newblock {\em Data mining and knowledge discovery}, 31:606--660.

\bibitem[Bagnall et~al., 2015]{bagnall2015time}
Bagnall, A., Lines, J., Hills, J., and Bostrom, A. (2015).
\newblock Time-series classification with cote: the collective of
  transformation-based ensembles.
\newblock {\em IEEE Transactions on Knowledge and Data Engineering},
  27(9):2522--2535.

\bibitem[Bartlett et~al., 2005]{bartlett2005local}
Bartlett, P.~L., Bousquet, O., and MENDELSON, S. (2005).
\newblock Local rademacher complexities.
\newblock {\em The Annals of Statistics}, 33(4):1497--1537.

\bibitem[Bartlett et~al., 2021]{bartlett2021deep}
Bartlett, P.~L., Montanari, A., and Rakhlin, A. (2021).
\newblock Deep learning: a statistical viewpoint.
\newblock {\em Acta numerica}, 30:87--201.

\bibitem[Cui et~al., 2016]{Cui2016MultiScaleCN}
Cui, Z., Chen, W., and Chen, Y. (2016).
\newblock Multi-scale convolutional neural networks for time series
  classification.
\newblock {\em ArXiv}, abs/1603.06995.

\bibitem[Cuturi and Blondel, 2017]{cuturi2017soft}
Cuturi, M. and Blondel, M. (2017).
\newblock Soft-dtw: a differentiable loss function for time-series.
\newblock In {\em International conference on machine learning}, pages
  894--903. PMLR.

\bibitem[Deng et~al., 2013]{DENG2013142}
Deng, H., Runger, G., Tuv, E., and Vladimir, M. (2013).
\newblock A time series forest for classification and feature extraction.
\newblock {\em Information Sciences}, 239:142--153.

\bibitem[Duan et~al., 2023]{duan2023fast}
Duan, C., Jiao, Y., Kang, L., Lu, X., and Yang, J.~Z. (2023).
\newblock {Fast excess risk rates via offset Rademacher complexity}.
\newblock In {\em International Conference on Machine Learning}, pages
  8697--8716. PMLR.

\bibitem[Eckmann et~al., 1995]{eckmann1995recurrence}
Eckmann, J.-P., Kamphorst, S.~O., Ruelle, D., et~al. (1995).
\newblock Recurrence plots of dynamical systems.
\newblock {\em World Scientific Series on Nonlinear Science Series A},
  16:441--446.

\bibitem[Geer, 2000]{geer2000empirical}
Geer, S.~A. (2000).
\newblock {\em Empirical Processes in M-estimation}, volume~6.
\newblock Cambridge university press.

\bibitem[Gin{\'e} and Nickl, 2021]{gine2021mathematical}
Gin{\'e}, E. and Nickl, R. (2021).
\newblock {\em Mathematical foundations of infinite-dimensional statistical
  models}.
\newblock Cambridge university press.

\bibitem[Gonz{\'a}lez-Rivera and Lin, 2013]{gonzalez2013constrained}
Gonz{\'a}lez-Rivera, G. and Lin, W. (2013).
\newblock Constrained regression for interval-valued data.
\newblock {\em Journal of Business \& Economic Statistics}, 31(4):473--490.

\bibitem[Han et~al., 2012]{han2012autoregressive}
Han, A., Hong, Y., and Wang, S. (2012).
\newblock Autoregressive conditional models for interval-valued time series
  data.
\newblock In {\em The 3rd International Conference on Singular Spectrum
  Analysis and Its Applications}, page~27.

\bibitem[Han et~al., 2016]{han2016vector}
Han, A., Hong, Y., Wang, S., and Yun, X. (2016).
\newblock A vector autoregressive moving average model for interval-valued time
  series data.
\newblock In {\em Essays in Honor of Aman Ullah}. Emerald Group Publishing
  Limited.

\bibitem[He and Tao, 2020]{he2020recent}
He, F. and Tao, D. (2020).
\newblock Recent advances in deep learning theory.
\newblock {\em arXiv preprint arXiv:2012.10931}.

\bibitem[He et~al., 2016]{he2016deep}
He, K., Zhang, X., Ren, S., and Sun, J. (2016).
\newblock Deep residual learning for image recognition.
\newblock In {\em Proceedings of the IEEE conference on computer vision and
  pattern recognition}, pages 770--778.

\bibitem[He et~al., 2021]{he2021forecasting}
He, Y., Han, A., Hong, Y., Sun, Y., and Wang, S. (2021).
\newblock Forecasting crude oil price intervals and return volatility via
  autoregressive conditional interval models.
\newblock {\em Econometric Reviews}, 40(6):584--606.

\bibitem[Hu and Qi, 2019]{DBLP:journals/corr/abs-1901-09891}
Hu, T. and Qi, H. (2019).
\newblock See better before looking closer: Weakly supervised data augmentation
  network for fine-grained visual classification.
\newblock {\em CoRR}, abs/1901.09891.

\bibitem[Ismail~Fawaz et~al., 2019]{ismail2019deep}
Ismail~Fawaz, H., Forestier, G., Weber, J., Idoumghar, L., and Muller, P.-A.
  (2019).
\newblock Deep learning for time series classification: a review.
\newblock {\em Data mining and knowledge discovery}, 33(4):917--963.

\bibitem[Jentzen et~al., 2023]{jentzen2023mathematical}
Jentzen, A., Kuckuck, B., and von Wurstemberger, P. (2023).
\newblock Mathematical introduction to deep learning: Methods, implementations,
  and theory.
\newblock {\em arXiv preprint arXiv:2310.20360}.

\bibitem[Koltchinskii, 2006]{koltchinskii2006local}
Koltchinskii, V. (2006).
\newblock Local rademacher complexities and oracle inequalities in risk
  minimization. discussion.
\newblock {\em Annals of statistics}, 34(6):2593--2706.

\bibitem[Liang et~al., 2015]{liang2015learning}
Liang, T., Rakhlin, A., and Sridharan, K. (2015).
\newblock {Learning with square loss: Localization through offset rademacher
  complexity}.
\newblock In {\em Conference on Learning Theory}, pages 1260--1285. PMLR.

\bibitem[Lines and Bagnall, 2015]{lines2015time}
Lines, J. and Bagnall, A. (2015).
\newblock Time series classification with ensembles of elastic distance
  measures.
\newblock {\em Data Mining and Knowledge Discovery}, 29(3):565--592.

\bibitem[Lu et~al., 2022]{lu2022forecasting}
Lu, Q., Sun, Y., Hong, Y., and Wang, S. (2022).
\newblock Forecasting interval-valued crude oil prices using asymmetric
  interval models.
\newblock {\em Quantitative Finance}, 22(11):2047--2061.

\bibitem[Lucas et~al., 2019]{lucas2019proximity}
Lucas, B., Shifaz, A., Pelletier, C., O’Neill, L., Zaidi, N., Goethals, B.,
  Petitjean, F., and Webb, G.~I. (2019).
\newblock Proximity forest: an effective and scalable distance-based classifier
  for time series.
\newblock {\em Data Mining and Knowledge Discovery}, 33(3):607--635.

\bibitem[Maia et~al., 2008]{maia2008forecasting}
Maia, A. L.~S., de~Carvalho, F. d.~A., and Ludermir, T.~B. (2008).
\newblock Forecasting models for interval-valued time series.
\newblock {\em Neurocomputing}, 71(16-18):3344--3352.

\bibitem[Munkres, 2018]{munkres2018elements}
Munkres, J.~R. (2018).
\newblock {\em Elements of algebraic topology}.
\newblock CRC press.

\bibitem[Nguyen and Ifrim, 2021]{Nguyen2021MrSQMFT}
Nguyen, T.~L. and Ifrim, G. (2021).
\newblock Mrsqm: Fast time series classification with symbolic representations.
\newblock {\em ArXiv}, abs/2109.01036.

\bibitem[Palumbo and Verde, 1999]{palumbo1999non}
Palumbo, F. and Verde, R. (1999).
\newblock Non-symmetrical factorial discriminant analysis for symbolic objects.
\newblock {\em Applied Stochastic Models in Business and Industry},
  15(4):419--427.

\bibitem[Qi et~al., 2020]{qi2020interval}
Qi, X., Guo, H., Artem, Z., and Wang, W. (2020).
\newblock An interval-valued data classification method based on the unified
  representation frame.
\newblock {\em IEEE Access}, 8:17002--17012.

\bibitem[Rasson and Lissoir, 2000]{rasson2000symbolic}
Rasson, J.-P. and Lissoir, S. (2000).
\newblock Symbolic kernel discriminant analysis.
\newblock {\em Computational Statistics}, 15(1):127--132.

\bibitem[Romano et~al., 2004]{ROMANO2004214}
Romano, M.~C., Thiel, M., Kurths, J., and {von Bloh}, W. (2004).
\newblock Multivariate recurrence plots.
\newblock {\em Physics Letters A}, 330(3):214--223.

\bibitem[Ruiz et~al., 2021]{ruiz2021great}
Ruiz, A.~P., Flynn, M., Large, J., Middlehurst, M., and Bagnall, A. (2021).
\newblock The great multivariate time series classification bake off: a review
  and experimental evaluation of recent algorithmic advances.
\newblock {\em Data Mining and Knowledge Discovery}, 35(2):401--449.

\bibitem[Sch{\"a}fer, 2015]{schafer2015boss}
Sch{\"a}fer, P. (2015).
\newblock The boss is concerned with time series classification in the presence
  of noise.
\newblock {\em Data Mining and Knowledge Discovery}, 29:1505--1530.

\bibitem[Sch{\"a}fer, 2016]{schafer2016scalable}
Sch{\"a}fer, P. (2016).
\newblock Scalable time series classification.
\newblock {\em Data Mining and Knowledge Discovery}, 30(5):1273--1298.

\bibitem[Sch{\"a}fer and Leser, 2017]{schafer2017fast}
Sch{\"a}fer, P. and Leser, U. (2017).
\newblock Fast and accurate time series classification with weasel.
\newblock In {\em Proceedings of the 2017 ACM on Conference on Information and
  Knowledge Management}, pages 637--646.

\bibitem[Sun et~al., 2018]{sun2018threshold}
Sun, Y., Han, A., Hong, Y., and Wang, S. (2018).
\newblock Threshold autoregressive models for interval-valued time series data.
\newblock {\em Journal of Econometrics}, 206(2):414--446.

\bibitem[Sun et~al., 2019]{sun2019asymmetric}
Sun, Y., Zhang, X., Hong, Y., and Wang, S. (2019).
\newblock Asymmetric pass-through of oil prices to gasoline prices with
  interval time series modelling.
\newblock {\em Energy Economics}, 78:165--173.

\bibitem[Sun et~al., 2022]{sun2022model}
Sun, Y., Zhang, X., Wan, A.~T., and Wang, S. (2022).
\newblock Model averaging for interval-valued data.
\newblock {\em European Journal of Operational Research}, 301(2):772--784.

\bibitem[Tian and Qin, 2024]{wan2023discriminant}
Tian, W. and Qin, Z. (2024).
\newblock Adaptive classification of interval-valued time series.
\newblock Unpublished manuscript.

\bibitem[Utkin et~al., 2016]{Utkin2016}
Utkin, L.~V., Chekh, A.~I., and Zhuk, Y.~A. (2016).
\newblock Binary classification {SVM}-based algorithms with interval-valued
  training data using triangular and epanechnikov kernels.
\newblock {\em Neural Networks}, 80:53--66.

\bibitem[Vaart and Wellner, 2023]{vaart2023empirical}
Vaart, A. v.~d. and Wellner, J.~A. (2023).
\newblock Empirical processes.
\newblock In {\em Weak Convergence and Empirical Processes: With Applications
  to Statistics}, pages 127--384. Springer.

\bibitem[Van~der Vaart, 2000]{van2000asymptotic}
Van~der Vaart, A.~W. (2000).
\newblock {\em Asymptotic statistics}, volume~3.
\newblock Cambridge university press.

\bibitem[Vershynin, 2018]{vershynin2018high}
Vershynin, R. (2018).
\newblock {\em High-dimensional probability: An introduction with applications
  in data science}, volume~47.
\newblock Cambridge university press.

\bibitem[Wang et~al., 2023]{wanginterval}
Wang, P., Gurmani, S.~H., Tao, Z., Liu, J., and Chen, H. (2023).
\newblock Interval time series forecasting: A systematic literature review.
\newblock {\em Journal of Forecasting}.

\bibitem[Wang et~al., 2017]{Wang2017}
Wang, Y., Cembrano, G., Puig, V., Urrea, M., Romera, J., Saporta, D., Valero,
  J., and Quevedo, J. (2017).
\newblock Optimal management of barcelona water distribution network using
  non-linear model predictive control.
\newblock {\em {IFAC}-{PapersOnLine}}, 50(1):5380--5385.

\bibitem[Wang and Oates, 2015]{wang2015encoding}
Wang, Z. and Oates, T. (2015).
\newblock Encoding time series as images for visual inspection and
  classification using tiled convolutional neural networks.
\newblock In {\em Workshops at the twenty-ninth AAAI conference on artificial
  intelligence}.

\bibitem[Wang et~al., 2016]{Wang2016TimeSC}
Wang, Z., Yan, W., and Oates, T. (2016).
\newblock Time series classification from scratch with deep neural networks: A
  strong baseline.
\newblock {\em 2017 International Joint Conference on Neural Networks (IJCNN)},
  pages 1578--1585.

\bibitem[Yang et~al., 2016]{yang2016analysis}
Yang, W., Han, A., Hong, Y., and Wang, S. (2016).
\newblock Analysis of crisis impact on crude oil prices: a new approach with
  interval time series modelling.
\newblock {\em Quantitative Finance}, 16(12):1917--1928.

\bibitem[Zhang, 2023]{zhang2023mathematical}
Zhang, T. (2023).
\newblock {\em Mathematical analysis of machine learning algorithms}.
\newblock Cambridge University Press.

\bibitem[Zhao and Itti, 2017]{2017shapeDTW}
Zhao, J. and Itti, L. (2017).
\newblock shapedtw: shape dynamic time warping.
\newblock {\em Pattern Recognition}, 74:171--184.

\end{thebibliography}
\bibliographystyle{apalike}

\end{document}